\documentclass{article}

\usepackage[utf8]{inputenc}
\usepackage[english]{babel}
\usepackage{csquotes}

\usepackage[letterpaper,top=2cm,bottom=2cm,left=3cm,right=3cm,marginparwidth=1.75cm]{geometry}
\usepackage{amsmath}
\usepackage{graphicx}
\usepackage{booktabs}
\usepackage{multicol}
\usepackage{multirow}
\usepackage[colorlinks=true, allcolors=blue]{hyperref}
\usepackage{amssymb}
\usepackage{amsfonts}
\usepackage{rotating} 
\usepackage{listings}
\usepackage{xcolor}
\usepackage{caption}
\usepackage{subcaption}
\usepackage{pdflscape}
\usepackage{textcomp}
\usepackage{xurl}
\usepackage[backend=biber, style=nature, url=true, doi=true]{biblatex}
\usepackage[T1]{fontenc}
\usepackage{threeparttable}
\usepackage{array}
\usepackage{makecell}

\definecolor{codegreen}{rgb}{0,0.6,0}
\definecolor{codegray}{rgb}{0.5,0.5,0.5}
\definecolor{codepurple}{rgb}{0.58,0,0.82}
\definecolor{backcolour}{rgb}{0.95,0.95,0.92}

\lstdefinestyle{mystyle}{
    backgroundcolor=\color{backcolour},   
    commentstyle=\color{codegreen},
    keywordstyle=\color{magenta},
    numberstyle=\tiny\color{codegray},
    stringstyle=\color{codepurple},
    basicstyle=\ttfamily\footnotesize,
    breakatwhitespace=false,         
    breaklines=true,                 
    captionpos=b,                    
    keepspaces=true,                 
    numbers=left,                    
    numbersep=5pt,                  
    showspaces=false,                
    showstringspaces=false,
    showtabs=false,                  
    tabsize=2
}
\definecolor{deepblue}{rgb}{0,0,0.5}
\definecolor{deepred}{rgb}{0.6,0,0}
\definecolor{deepgreen}{rgb}{0,0.5,0}

\lstnewenvironment{python}[1][] 
{\lstset{language=Python, #1}} 
{}

\title{A Self-Evolving Agentic System for Automated Generation and Execution of Biological Protocols}
\author{Yankai Jiang$^{1\ast \dagger}$, Weiting Tang$^{2\ast}$, Haoran Sun$^{1\ast}$, Zhenyu Tang$^{1\ast}$, Yuejie Hou$^{2\ast}$, \\ Yingnan Han$^{1}$, Rubo Wang$^{1}$, Yueyuxiao Yang$^{2}$, Cheng Liang$^{2}$, \\ Lilong Wang$^{1}$, Wenjie Lou$^{1}$, Xiaosong Wang$^{1\dagger}$, Lei Bai$^{1\dagger}$, Meng Yang$^{1,2\dagger}$}

\begin{document}
\maketitle

\vspace{-3em}
\begin{center}
   \begin{tabular}{c}
        $^{1}$Shanghai Artificial Intelligence Laboratory, Shanghai, China \\ 
        $^{2}$Genoria AI, Shenzhen, China \\[5pt]
        $\ast$ These authors contributed equally \\
       $\dagger$ Corresponding authors \\
        Corresponding author(s): jiangyankai@pjlab.org.cn, wangxiaosong@pjlab.org.cn, \\ bailei@pjlab.org.cn, yangmeng@genoria.ai
    \end{tabular}
\end{center}

\begin{abstract}
Autonomous wet-lab experimentation requires more than plausible protocol text: biological intent, quantitative procedures, device constraints and experimental feedback must remain aligned from protocol and SOP design to code and physical execution. We developed ProtoPilot, a self-evolving multi-agent system, together with an expert-grounded benchmark and evaluation framework for testing this conversion as an experimental automation problem. The framework spans 294 synthetic-biology and molecular-biology tasks derived from 98 gold-standard protocols, wet-lab expert rubrics, device-level validity gates and real experimental tests. ProtoPilot incorporates  layer-wise verifiability, multi-agent orchestration and a runtime-updated skill library to generate protocols, expand SOPs, synthesize SDK-compliant code and revise workflows from wet-lab feedback. It achieved a Top$@$3 expert-preference rate of $90.2\%$, an overall protocol-to-code gate pass rate of $96.6\%$ and an Opentrons pass rate of $88.2\%$, compared with $32.4\%$ for OpenTrons-AI. Wet-lab validation produced interpretable readouts, Sanger-confirmed products and feedback-corrected PCA-assembled DNA targets, establishing a verifiable route to autonomous experimentation. Together, these results show that the evaluation framework captures execution-relevant requirements for autonomous wet-lab automation, and that ProtoPilot can meet them by converting protocol and code generation into validated execution and feedback-guided revision.

\end{abstract}

\section{Introduction} 
AI agents for the life sciences are crossing the boundary from digital reasoning to physical experimentation. Recent systems show that language-model agents can organize literature evidence, generate scientific hypotheses, support experimental design and participate in focused wet-lab workflows~\cite{gottweis2026accelerating,swanson2025virtual,gao2024empowering,ghareeb2026multi,wang2026accelerating}. Robot Scientist systems, self-driving laboratories and automated cloud-lab studies further show that experimental platforms can support controlled biochemical operations, robotic execution and iterative optimization~\cite{king2009automation,rapp2024self,smith2026using}. Yet the handoff from biological intent to physical experimentation remains fragile. A scientific idea becomes experimentally actionable only when the intended biology is translated into samples, reagents, parameters, procedures, device actions, quality-control checks, readouts and revision decisions that remain valid on real instruments and under real experimental outcomes.

\begin{figure}[htbp]
  \centering
  \includegraphics[width=0.95\textwidth]{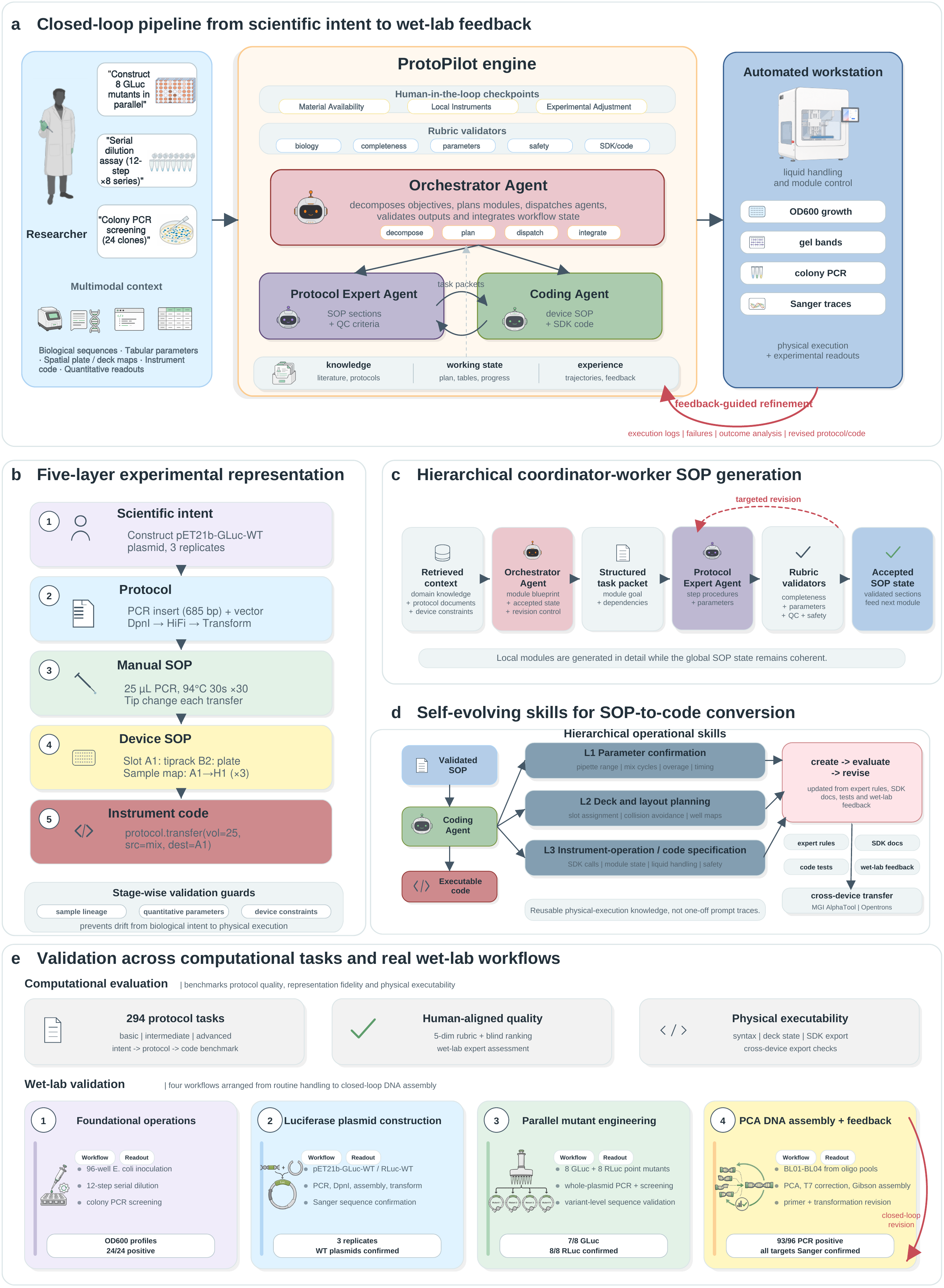}
  \caption{%
    \textbf{Overview of ProtoPilot.
    }
    \textbf{(a)}~Closed-loop pipeline from scientific intent to wet-lab feedback.
    \textbf{(b)}~Five-layer experimental representation, from scientific intent
    through protocol, manual SOP and device SOP to instrument code, with
    stage-wise guards preserving consistency across layers.
    \textbf{(c)}~Hierarchical coordinator-worker mechanism for SOP generation, in which the
    Orchestrator gates step procedures from the Protocol Expert Agent through
    rubric validators and returns failing modules for targeted revision.
    \textbf{(d)}~Self-evolving skills for SOP-to-code conversion, in which the
    Coding Agent applies three hierarchical skill classes refined through a
    create, evaluate and revise loop to yield reusable, cross-device execution
    knowledge.
    \textbf{(e)}~Validation across 294 computational tasks and four wet-lab
    workflows of increasing complexity.
  }
  \label{fig:fig1}
\end{figure}

Designing a protocol from natural-language scientific intent is a long-context generation task that demands domain-specific expertise, long-range memory, and accumulated practical experience~\cite{jin2025biolab,sun2025unleashing,li2024loogle}. Moreover, a protocol draft is an intermediate representation rather than an executable experiment~\cite{ananthanarayanan2010biocoder,gao2025autonomous}. During wet-lab automation, scientific rationale, sample and reagent states, quantitative parameters, operation order, device constraints and quality-control readouts must remain aligned as a natural-language goal becomes a protocol, SOP, device workflow, executable code and feedback-guided revision (Fig.~\ref{fig:fig1}a,b). Different layers resolve different ambiguities. A protocol must express biological logic, sample lineage, timing and quality-control structure. An SOP must ground that logic in feasible volumes, concentrations, labware, incubation conditions and acceptance criteria. Device code must bind the procedure to deck layout, well mapping, liquid-handling actions, timing control and vendor-specific SDK commands. Failures of alignment across these layers can make a workflow difficult to execute, reproduce or repair after an unsuccessful run~\cite{bartley2023building,anhel2023laboratory}.

Recent work has advanced important parts of this transformation. Co-Scientist, Robin and the Virtual Lab establish language-model agents as tools for scientific reasoning, hypothesis generation and experimental design~\cite{gottweis2026accelerating,swanson2025virtual,ghareeb2026multi}. PrimeGen connects agentic design with liquid-handling code, anomaly monitoring and wet-lab readouts in targeted biological workflows~\cite{wang2026accelerating}. Thoth, BioProAgent and LabscriptAI improve protocol generation, constrained experimental planning and protocol-to-script conversion~\cite{sun2025unleashing,liu2026bioproagent,gao2025autonomous}. BPL, LabOP, BioCoder and LAP provide more explicit representations for protocol structure, type constraints, state tracking, automation scripts and compiler-facing diagnostics~\cite{song2026towards,ananthanarayanan2010biocoder,bartley2023building,anhel2023laboratory}. Together, these studies show that AI-enabled laboratory execution now has concrete technical precedents. They also sharpen a methodological requirement: autonomous wet-lab agents should be evaluated not only by the plausibility of generated protocol text, but by whether biological intent remains intact through executable, testable and revisable experimental workflows.

We addressed this requirement by constructing BioLab bench, an expert-grounded benchmark and evaluation framework for autonomous wet‑lab experimentation. The benchmark combines diverse synthetic-biology and molecular-biology scenarios, gold-standard protocol data, wet-lab expert experience, task-specific metrics, rubric-based validators, device-level validity gates and real experimental tests. Rather than treating protocol generation as an isolated text task, the framework evaluates whether generated workflows satisfy biological intent, preserve methodological consistency, instantiate sound parameters and remain procedurally complete. It further tests whether those workflows can be translated into SDK-exportable, instrument-compatible command objects, and whether they produce interpretable biological readouts that can support revision after failed or suboptimal steps. In this way, the benchmark turns expert wet-lab judgment into measurable requirements for experimental automation, and underscores the gap between the capabilities of existing general‑purpose or specialized AI agents and those required for truly autonomous wet‑lab experimentation.

Motivated by this gap, we developed ProtoPilot, a self-evolving multi-agent system for layer-wise verifiable experimental automation. ProtoPilot is organized around two mechanisms. The first is hierarchical multi-agent collaboration. An Orchestrator Agent maintains workflow state while specialized agents generate protocols, expand SOPs, map procedures to devices, synthesize SDK-gated code, validate intermediate outputs and incorporate wet-lab feedback. Each transition produces an inspectable object, from protocol and SOP to device workflow, command object and readout. The second mechanism is runtime agent-skill learning. A multilayer agentic memory and self-evolving skill library encode reusable operational knowledge from domain knowledge, protocol context, SDK information, code verification, task history, expert feedback and experimental outcomes. These skills support parameter confirmation, deck and sample-layout planning, and instrument-operation code specification, enabling underspecified experimental steps to be grounded as SDK-compliant workflows and adapted to new tasks or instruments. Rubric-based validators connect these mechanisms to the benchmark criteria by checking biological plausibility, procedural completeness, parameter consistency, safety, protocol-code alignment and instrument compatibility.

Applying this framework, we evaluated ProtoPilot across protocol generation, protocol-to-code translation, cross-platform robotic execution and wet-lab validation. The benchmark contains 294 synthetic-biology and molecular-biology tasks derived from 98 gold-standard protocols across increasing levels of experimental complexity. ProtoPilot-generated protocols received stronger blind preferences from wet-lab experts, reaching a Top@3 preference rate of $90.2\%$. In protocol-to-code tasks, generated workflows passed syntax, parameter-alignment and MGI AlphaTool SDK export gates with an overall gate pass rate of $96.6\%$.
Additional Opentrons evaluation showed a ProtoPilot pass rate of $88.2\%$, compared with $32.4\%$ for OpenTrons-AI, indicating transfer beyond a single device platform. 
We further validate ProtoPilot across wet-lab workflows of increasing complexity, including routine liquid handling and clone screening, luciferase plasmid construction, whole-plasmid mutagenesis and PCA-based de novo DNA assembly (Fig.~\ref{fig:fig1}e). These experiments yielded reproducible 96-well bacterial inoculation, ordered tenfold dilution profiles, positive colony-PCR products in all 24 tested clones, Sanger-confirmed GLuc-WT and RLuc-WT plasmids, confirmed mutants for seven of eight GLuc and all eight RLuc designs, and sequence-confirmed PCA-assembled EGFP and A501\_Ec.CPS-derived DNA targets after feedback-guided protocol revision. In the PCA experiment, ProtoPilot revised the protocol after failures in vector amplification and transformation, leading to sequence-confirmed assembled DNA targets in subsequent experiments. Together, the comprehensive evaluation demonstrate that ProtoPilot can 
provide a paired evaluation-and-execution framework for converting biological intent into executable, verifiable and feedback-correctable wet-lab workflows.


\section{Results}

\subsection{ProtoPilot Overview: A self-evolving multi-agent system for biological experiment automation}
ProtoPilot was designed to operationalize end-to-end wet-lab automation within a single stateful agent loop. Starting from a free-form natural-language description of an experimental objective, ProtoPilot progressively transforms user intent into a validated SOP, an instrument-executable program and a physically executed workflow, and then uses execution outcomes to guide subsequent refinement (Fig. 1a). The system can operate autonomously, while also supporting human-in-the-loop checkpoints when user confirmation or local laboratory constraints are present.

At the core of ProtoPilot is an Orchestrator Agent that maintains a structured workflow state. The Orchestrator decomposes the user objective into actionable subtasks, plans the workflow and coordinates two specialized agents: a Protocol Expert Agent trained with reinforcement learning for step-level SOP generation, and a Coding Agent equipped with a self-evolving skill library for translating validated protocols into instrument-executable code. Domain knowledge, prior task experience and the current workflow state are organized in a multilayer agentic memory bank. This memory includes a knowledge base of life-science domain knowledge, literature and retrievable protocol documents; a working memory that records the plan, constraints, intermediate outputs and progress of the current task; and a task-experience base that stores completed trajectories, user preferences, validation results, execution traces and feedback. The Orchestrator retrieves from and updates this memory throughout the workflow to support planning, consistency checking and inter-agent coordination. Across the loop, rubric-based validators evaluate generated SOPs and code for biological plausibility, procedural completeness, parameter consistency, safety, protocol--code alignment and compatibility with instrument and SDK constraints.

Beyond the default autonomous execution mode, ProtoPilot supports real-time human-in-the-loop interaction. At key decision points, the Orchestrator Agent records the current generation state and queries the user to confirm whether the required materials are available, whether the intended instruments can support the proposed operations and whether experimental adjustments are needed. After execution, experimental observations, quantitative readouts and outcome analyses are returned to the Orchestrator, which updates the workflow state, replans the next iteration and invokes the relevant expert agent to refine the protocol or selected experimental steps. This design allows ProtoPilot to connect planning, validation, execution and empirical feedback without treating each experimental attempt as an isolated prompt-response task.

ProtoPilot implements two core mechanisms to support this workflow. The first is a hierarchical, coarse-to-fine coordinator--worker mechanism for SOP construction. Rather than generating an entire SOP in a single pass, the Orchestrator first builds a module-level plan and global constraint set from the user objective, retrieved domain knowledge, prior task experience and relevant protocol literature. For each module, it provides the Protocol Expert Agent with the module goal, accepted upstream content, expected downstream dependencies, user and device constraints and accumulated SOP state. The expert agent then returns step-level procedures and candidate quantitative parameters. The Orchestrator evaluates the output against rigorously specified rubric-based criteria before integrating accepted content into the evolving SOP and planning the next module. This recurrent interaction enables local step-level detail while preserving global workflow state and long-range procedural coherence.

The second mechanism is self-evolving agent--skill learning for SOP-to-code conversion. The Coding Agent does not treat executable-code generation as a direct translation from protocol text to API calls. Instead, it uses a runtime-updated skill library initialized and expanded from expert-derived operational knowledge, including practical rules for material handling, device use, failure avoidance and biological constraint management. These skills are organized into three hierarchical classes: parameter-confirmation skills, deck- and sample-layout planning skills, and instrument-operation/code-specification skills. During conversion, parameter-confirmation skills resolve underspecified or qualitative SOP descriptions into executable values; layout-planning skills determine physical deck configurations and well-position maps; and operation/code-specification skills generate step-level operation descriptions together with SDK-compliant executable code. During runtime use and testing, candidate skills are iteratively improved through creation--evaluation--revision loops, in which skills are created from expert rules, protocol context, SDK information and prior task trajectories; evaluated using rubric-based validators, code checks and execution feedback; and revised into more reliable reusable routines. By encoding stable expert operational principles rather than surface features of a single protocol or SDK, this skill layer provides a transferable basis for protocol--code consistency and adaptation across experimental tasks, device models and vendor-specific software interfaces.

Overall, ProtoPilot treats laboratory automation as a stateful, feedback-driven process rather than a set of independent generation tasks. By integrating hierarchical agent orchestration, multilayer memory, rubric-based validation, runtime skill learning, optional human oversight and automated workstation execution, ProtoPilot converts natural-language experimental goals into validated SOPs and SDK-compliant executable programs, orchestrates their execution on physical instruments and updates subsequent designs from experimental outcomes within a unified closed loop.

\begin{figure}[htbp]
  \centering
  \includegraphics[width=0.97\textwidth]{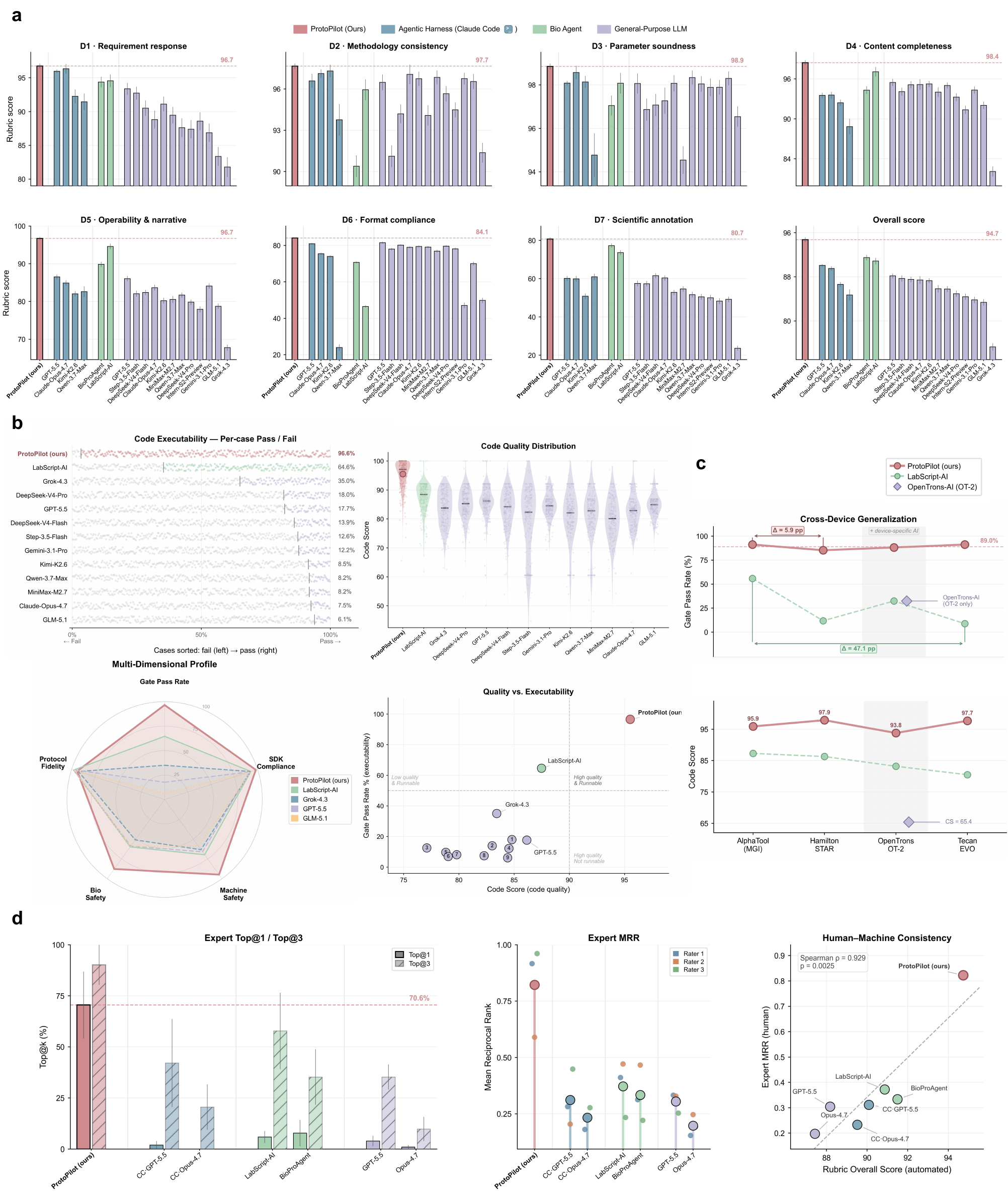}
  \caption{%
    \textbf{ProtoPilot evaluation across protocol quality, code executability,
    cross-device generalization, and expert preference.}
    \textbf{(a)}~Rubric-based protocol quality scores across 294 test cases
    spanning three difficulty tiers (L1--L3) and seven criteria (D1--D7),
    shown per system (mean~$\pm$~s.e.m.).
    \textbf{(b)}~Protocol$\rightarrow$Code evaluation across 13 systems
    ($n = 294$ cases per system), comprising per-case executability strips
    (gate pass/fail); Code Score distributions; a multi-dimensional capability
    radar covering Gate Rate, SDK Compliance, Machine Safety, Bio Safety, and
    Protocol Fidelity; and a quality--executability scatter (Code Score,
    $x$-axis; Gate Rate, $y$-axis).
    \textbf{(c)}~Gate Pass Rate (upper) and Protocol Score (lower) across four
    laboratory automation platforms (AlphaTool MGI, Hamilton STAR, OpenTrons
    OT-2, Tecan EVO; $n = 34$ tasks per device) for ProtoPilot, LabScript-AI,
    and OpenTrons-AI (OT-2 only).
    \textbf{(d)}~Blind expert preference evaluation across 7 systems and 34
    queries rated independently by 3 wet-lab scientists.
    Grouped bars show Top@1 (solid fill) and Top@3 (hatched fill) preference
    rates per system; lollipop markers show mean reciprocal rank (MRR) with
    individual rater values as dots; scatter shows the Spearman correlation
    between automated rubric score ($x$-axis) and expert MRR ($y$-axis) at
    the system level ($n = 7$).
    Error bars, s.e.m.
  }
  \label{fig:fig2}
\end{figure}




\subsection{ProtoPilot achieves superior protocol quality across complexity levels, corroborated by blind expert evaluation}

We evaluated ProtoPilot against all comparison systems on 294 protocol generation tasks spanning three difficulty levels (L1--L3) using a seven-dimension rubric (D1--D7; see Technical Validation). ProtoPilot achieved an overall rubric score of 94.7~$\pm$~0.4 (mean~$\pm$~s.e.m.), outperforming the next-best system, BioProAgent, by 3.2 points and all general-purpose LLMs by more than 6 points (Fig.~\ref{fig:fig2}a). Performance remained consistently high across all difficulty tiers: 96.5~$\pm$~1.1 at L1 ($n = 15$, basic single-reagent workflows), 95.1~$\pm$~0.5 at L2 ($n = 165$, intermediate multi-reagent protocols), and 94.0~$\pm$~0.8 at L3 ($n = 114$, complex multi-stage workflows; Extended Data Fig.~\ref{fig:app_fig2a}). The 2.5-point decline from L1 to L3 was substantially smaller than the corresponding declines observed in agentic harness systems, confirming that ProtoPilot's quality advantage is not only preserved but widened under increasing domain-specific complexity.

Dimension-level analysis showed uniformly strong performance across the rubric profile. On the three core scientific quality dimensions, requirement response (D1: 96.7), methodology consistency (D2: 97.7), and parameter soundness (D3: 98.9) each reached near-maximal levels, confirming that ProtoPilot accurately addresses user intent, adheres to reference-document methodology, and correctly specifies quantitative parameters across protocol types. Operability and narrative coherence (D5: 96.7) showed a clear advantage over harness-augmented systems (82.0--86.5), a gap reflecting ProtoPilot's structured section transitions and step-level operational specificity. Scientific annotation depth (D7: 80.7) exhibited the largest separation from harness-augmented systems (50.8--61.0) and general-purpose LLMs (27.3--63.9), a differentiation attributable to the domain-specialized training underlying Thoth, ProtoPilot's dedicated protocol generation model.

The automated rubric was validated by a blind preference evaluation in which three independent wet-lab scientists ranked outputs from seven representative systems across 34 queries, without knowledge of system identity (see Technical Validation). ProtoPilot was ranked first in 70.6~$\pm$~16.4\% of queries and appeared within the top three in 90.2~$\pm$~9.8\% of queries, both substantially exceeding all comparison systems (BioProAgent: 7.8~$\pm$~6.4\% top-1; LabScript-AI: 57.8~$\pm$~18.7\% top-3; Fig.~\ref{fig:fig2}d), with consistent leads maintained across all difficulty tiers (Extended Data Fig.~\ref{fig:app_d1}). ProtoPilot's MRR was 0.82~$\pm$~0.12, more than twice that of the second-ranked system (LabScript-AI: 0.37~$\pm$~0.07), and head-to-head preference rates against individual systems ranged from 0.78 to 0.92 (Extended Data Fig.~\ref{fig:app_d4}). Inter-rater agreement across the three experts was moderate (overall Kendall's $W = 0.495$; Extended Data Fig.~\ref{fig:app_d3}), consistent with the inherent subjectivity of protocol quality ranking. Automated rubric scores and expert MRR were strongly correlated across all seven systems (Spearman $\rho = 0.929$, $p = 0.003$), confirming the rubric as a reliable proxy for independent expert judgement.

Together, these results establish that ProtoPilot consistently generates protocols of superior quality across all complexity levels, supported by both the highest automated rubric score among all evaluated systems and a top-1 expert preference rate of 70.6\% in blind evaluations. The strong concordance between automated and human assessment validates rubric-based scoring as a reliable and scalable measure of protocol quality for biological research workflows.

\subsection{ProtoPilot generates safe, executable instrument code with robust cross-device generalizability}

Beyond protocol quality, we assessed the safety and executability of instrument code generated by ProtoPilot and 12 comparison systems across all 294 protocol-to-code translation tasks using a six-metric Code Score and a three-layer gate check (Fig.~\ref{fig:fig2}b\footnote{Numbered systems in the bottom-right panel of \textbf{(b)}:
    1~DeepSeek-V4-Pro; 2~DeepSeek-V4-Flash; 3~Step-3.5-Flash; 4~Gemini-3.1-Pro;
    5~Kimi-K2.6; 6~MiniMax-M2.7; 7~Qwen-3.7-Max; 8~Claude-Opus-4.7; 9~GLM-5.1.}). 
ProtoPilot achieved a Protocol Score of 95.5, outperforming the next-best system (LabScript-AI, 87.5) by 8.0 points, with markedly lower variance across all tasks. Safety dimensions showed the largest margins: machine safety reached 94.8\% and bio safety 87.8\%, exceeding LabScript-AI by 25.1 and 28.0 percentage points, respectively. SDK compliance was similarly elevated (98.2\% versus 87.5--94.1\% across baselines; Extended Data Fig.~\ref{fig:app_b2}). Per-difficulty Code Score distributions confirm that ProtoPilot maintains a narrow, high-scoring profile at all complexity levels, whereas general-purpose LLMs exhibit substantially wider variance and lower medians at L3 (Extended Data Fig.~\ref{fig:app_b3}).

Executability was assessed through a three-layer gate check covering Python syntax validation, parameter alignment against the device SOP, and an SDK-level export that verified slot conflicts, pipetting volume bounds, and labware compatibility. ProtoPilot achieved an overall gate pass rate of 96.6\% across all 294 tasks, compared with a maximum of 35.0\% among general-purpose LLMs (Grok-4.3) and 64.6\% for LabScript-AI. Gate pass rates stratified by difficulty tier were 93.3\% at L1, 97.0\% at L2, and 96.3\% at L3 (Extended Data Fig.~\ref{fig:app_b1}). A joint analysis of Protocol Score and gate pass rate placed ProtoPilot alone in the high-quality, high-executability quadrant of this performance space (Fig.~\ref{fig:fig2}b): general-purpose LLMs achieved Protocol Scores of 77--86 yet gate pass rates below 35\%, demonstrating that surface-level code quality does not guarantee physical executability without coherent cross-layer alignment.

Cross-device generalizability was assessed by deploying ProtoPilot and LabScript-AI on four laboratory automation platforms (AlphaTool MGI, Hamilton STAR, OpenTrons OT-2, and Tecan EVO; $n = 34$ held-out tasks per device), with OpenTrons-AI additionally evaluated on OT-2 as a device-specific reference (Fig.~\ref{fig:fig2}c). ProtoPilot achieved gate pass rates of 91.2\%, 85.3\%, 88.2\%, and 91.2\% across the four platforms (a cross-device spread of 5.9 percentage points), whereas LabScript-AI ranged from 8.8\% to 55.9\% (a spread of 47.1 percentage points), revealing severe brittleness to platform-specific SDK requirements. On OT-2, where OpenTrons-AI (a purpose-built, single-platform system) was additionally benchmarked, ProtoPilot's gate pass rate of 88.2\% substantially exceeded both LabScript-AI (32.4\%) and OpenTrons-AI (32.4\%). Protocol Score followed an analogous pattern: ProtoPilot ranged from 93.8 to 97.9 across devices, while LabScript-AI ranged from 80.5 to 87.3 and OpenTrons-AI scored only 65.4 on its native platform.

Collectively, these results demonstrate that ProtoPilot is the only evaluated system to simultaneously achieve high code quality, near-complete executability, and stable cross-device performance, establishing it as a broadly deployable foundation for robotic wet-lab automation.

\subsection{ProtoPilot demonstrates reliable wet-lab performance across experiments of increasing complexity}

\subsubsection{ProtoPilot automates foundational wet-lab operations}

\begin{figure}[htbp]
  \centering
  \includegraphics[width=0.95\textwidth]{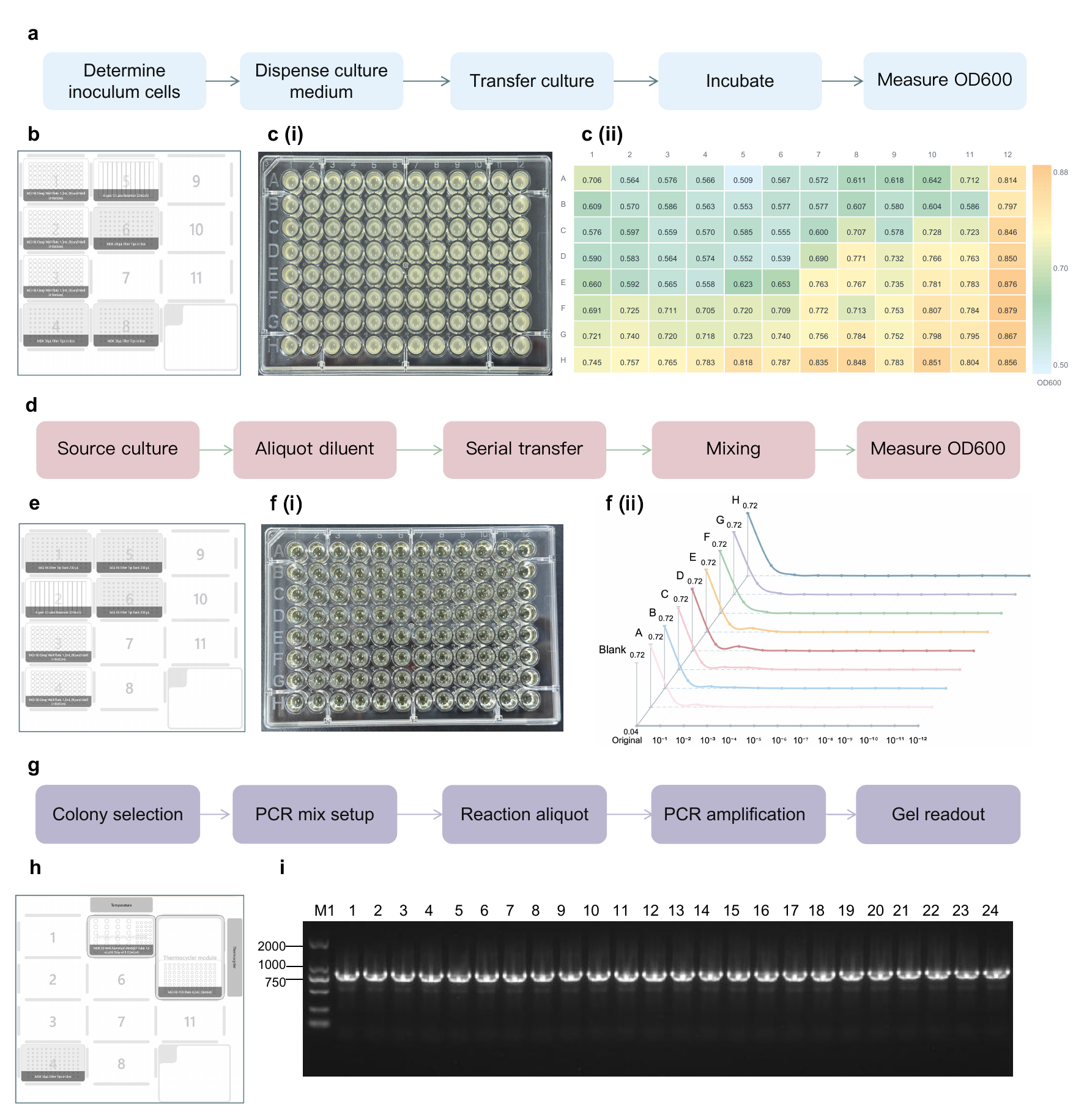}
  \caption{%
    \textbf{ProtoPilot automates three independent foundational wet-lab operations.}
    \textbf{(a)}~Workflow diagram for E. coli culture inoculation. 
    \textbf{(b)}~Automation deck layout for the E. coli culture inoculation experiment.
    \textbf{(c)}~Results of 96-well E. coli inoculation, including a representative culture plate after incubation (i) and an OD600 heatmap of sample wells (ii); heatmap colors indicate OD600 absorbance levels across wells. 
    \textbf{(d)}~Workflow diagram for serial dilution of E. coli liquid culture. 
    \textbf{(e)}~Automation deck layout for the serial dilution experiment. 
    \textbf{(f)}~Serial dilution results, including a representative dilution plate (i) and OD600 profiles across dilution levels (ii); the curves show the decrease in absorbance across tenfold serial dilutions.
    \textbf{(g)}~Workflow diagram for colony PCR screening.
    \textbf{(h)}~Automation deck layout for colony PCR screening. 
    \textbf{(i)}~Agarose gel electrophoresis of colony PCR products from 24 tested clones; M1 denotes DL2000 DNA marker, lanes 1-24 denote tested clones, and the target band indicates amplification of the expected fragment.
  }
  \label{fig:fig3}
\end{figure}

We first evaluated ProtoPilot on a set of high-frequency foundational wet-lab operations: \textit{E.~coli} culture inoculation, serial dilution, and colony PCR screening. These workflows represent routine steps in strain cultivation, sample dilution, and preliminary clone identification. Although individually straightforward, each task requires repetitive liquid handling, parallel processing across multiwell formats, and accurate transfer volumes, making them well-suited for assessing autonomous execution of routine laboratory procedures. After the user specified each task in natural language, ProtoPilot planned the experimental procedure and generated executable automation workflows covering source culture sampling, medium or diluent dispensing, culture transfer, mixing, and OD\textsubscript{600} measurement. Workflow diagrams summarize the major steps for the three independent experiments (Fig.~\ref{fig:fig3}a,d,g), and the corresponding automation layouts show representative deck configurations illustrating how each task was converted into an executable physical setup (Fig.~\ref{fig:fig3}b,e,h).

In the culture inoculation experiment, all 96 sample wells showed visible bacterial turbidity after incubation at 37\,°C with shaking, whereas blank control wells remained clear (Fig.~\ref{fig:fig3}c(i)
). The OD\textsubscript{600} heatmap showed that sample-well readings ranged from approximately 0.51 to 0.88, indicating that the automated workflow reproducibly supported multiwell culture inoculation and growth readout (Fig.~\ref{fig:fig3}c(ii)).

In the serial dilution experiment, bacterial culture with an initial OD\textsubscript{600} of approximately 0.71 was used to generate eight independent 12-step, tenfold dilution series. The representative plate image showed successful configuration of the dilution series in the multiwell format (Fig.~\ref{fig:fig3}f(i)). OD\textsubscript{600} profiles showed an ordered decrease in absorbance with increasing dilution, with the $10^{-1}$ dilution at approximately 0.13 and higher dilution levels approaching the blank background of approximately 0.039, with no detectable contamination across series (Fig.~\ref{fig:fig3}f(ii) and Extended Data Fig. 16b).

For colony PCR, ProtoPilot generated a workflow covering colony picking, PCR master-mix setup, reaction aliquoting, amplification, and gel readout (Fig.~\ref{fig:fig3}g), with PCR mix dispensing shown in Extended Data Fig. 16c. Agarose gel electrophoresis confirmed that all 24 tested clones produced clear, specific amplicons at the expected 868\,bp size, with no failed amplification or nonspecific bands (Fig.~\ref{fig:fig3}i), demonstrating that ProtoPilot can support sample handling, reaction setup, and product validation in basic clone-screening workflows.

Together, these experiments demonstrate that ProtoPilot can translate natural-language requests for common foundational wet-lab procedures into executable automation workflows, producing stable and experimentally verifiable readouts across diverse routine tasks.

\subsubsection{ProtoPilot supports luciferase plasmid construction and mutant engineering}

\noindent \textbf{2.4.2.1 Construction of pET21b-GLuc-WT and pET21b-RLuc-WT plasmids}

\begin{figure}[htbp]
  \centering
  \includegraphics[width=0.95\textwidth]{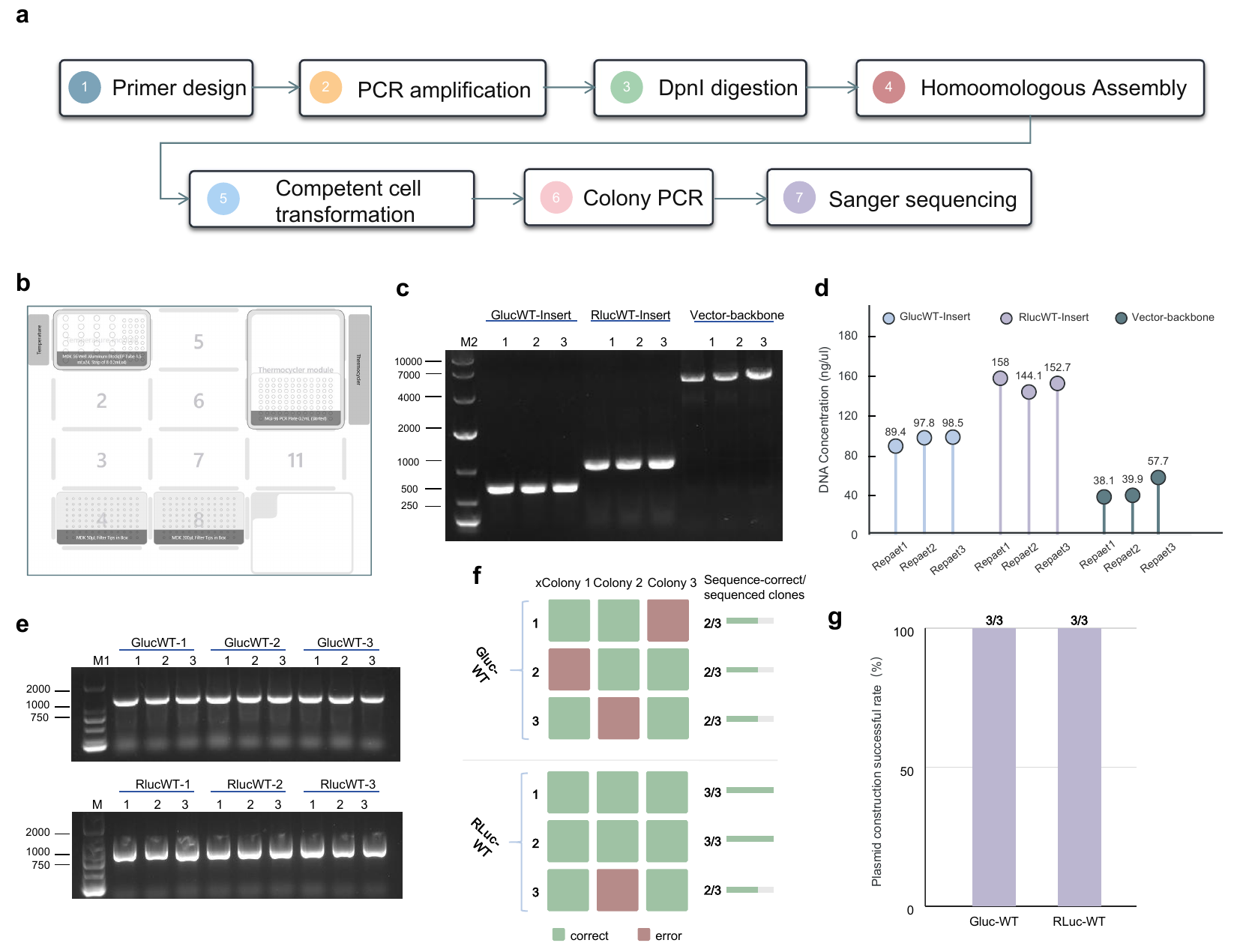}
  \caption{%
    \textbf{ProtoPilot supports construction of pET21b-GLuc-WT and pET21b-RLuc-WT plasmids.}
    \textbf{(a)}~Workflow diagram for wild-type luciferase plasmid construction.
    \textbf{(b)}~Automation deck layout for the wild-type luciferase plasmid construction experiment.
    \textbf{(c)}~Agarose gel electrophoresis of PCR products for the GLuc-WT insert, RLuc-WT insert and pET-21b(+) vector backbone; M2 denotes DL10000 DNA marker, and lanes 1-3 denote three independent replicates.
    \textbf{(d)}~DNA concentrations of purified PCR products, showing recovered concentrations of the GLuc-WT insert, RLuc-WT insert and vector backbone across three independent replicates.
    \textbf{(e)}~Colony PCR screening of candidate clones after transformation; M1 denotes DL2000 DNA marker, and three candidate clones were tested for each of three independent construction replicates per construct.
    \textbf{(f)}~Sanger sequencing confirmation matrix for GLuc-WT and RLuc-WT candidate clones; green indicates sequence-confirmed correct clones, red indicates unconfirmed or erroneous clones, and the horizontal bars on the right indicate the number of sequence-correct clones in each replicate.
    \textbf{(g)}~Final construction success rates for GLuc-WT and RLuc-WT plasmids, indicating whether at least one Sanger-confirmed correct plasmid clone was obtained for each target construct.
  }
  \label{fig:fig4}
\end{figure}

After validating foundational wet-lab operations, we next evaluated pET21b-GLuc-WT and pET21b-RLuc-WT plasmid construction as representative multi-step molecular cloning tasks. Unlike the foundational operations above, wild-type plasmid construction requires coordinated execution of insert amplification, vector-backbone preparation, DpnI digestion, homologous recombination assembly, competent-cell transformation, colony PCR screening, and Sanger confirmation, placing greater demands on step-to-step continuity, sample tracking, and quality control of intermediate products (Fig.~\ref{fig:fig4}a).

Upon receiving the user's initial request, ProtoPilot reasoned over the construction objectives and experimental route, then used multi-turn interaction to clarify the insert sequences, vector backbone, validation strategy, and available laboratory reagents, consumables, and equipment. After the experimental plan was confirmed, ProtoPilot generated an executable automation workflow supporting PCR amplification of the GLuc-WT and RLuc-WT inserts and the pET-21b(+) vector backbone, followed by assembly setup, transformation, and clone validation. The corresponding automation layout shows a representative deck configuration for the workflow on the liquid-handling platform (Fig.~\ref{fig:fig4}b).

Agarose gel electrophoresis of three independent replicates produced clear single bands at the expected sizes for the GLuc-WT insert (685\,bp), RLuc-WT insert (1,060\,bp), and pET-21b(+) vector backbone (5,403\,bp), with no nonspecific amplification (Fig.~\ref{fig:fig4}c). Quantification of the purified PCR products showed that the GLuc-WT insert concentrations across the three replicates were 89.4, 97.8, and 98.5\,ng/\textmu L; the RLuc-WT insert concentrations were 158.0, 144.1, and 152.7\,ng/\textmu L; and the pET-21b(+) vector-backbone concentrations were 38.1, 39.9, and 57.7\,ng/\textmu L, respectively, all sufficient for downstream homologous recombination assembly (Fig.~\ref{fig:fig4}d).

Candidate clones obtained after transformation were screened by colony PCR. For both GLuc-WT and RLuc-WT, all three colonies picked from each of the three independent construction replicates were PCR-positive (Fig.~\ref{fig:fig4}e). Subsequent Sanger sequencing confirmed two of the three candidate clones per replicate for GLuc-WT, and three, three, and two candidate clones across the three replicates for RLuc-WT (Fig.~\ref{fig:fig4}f). Both targets thus yielded Sanger-confirmed correct plasmid clones, with a final construction success rate of 100\% for both GLuc-WT and RLuc-WT (Fig.~\ref{fig:fig4}g). These results demonstrate that ProtoPilot can support a complete plasmid construction workflow spanning fragment preparation, vector processing, assembly setup, and positive-clone confirmation, while maintaining traceable execution across a multi-step molecular cloning task.

\noindent \textbf{2.4.2.2 Whole-plasmid mutagenesis of GLuc and RLuc}

\begin{figure}[htbp]
  \centering
  \includegraphics[width=0.95\textwidth]{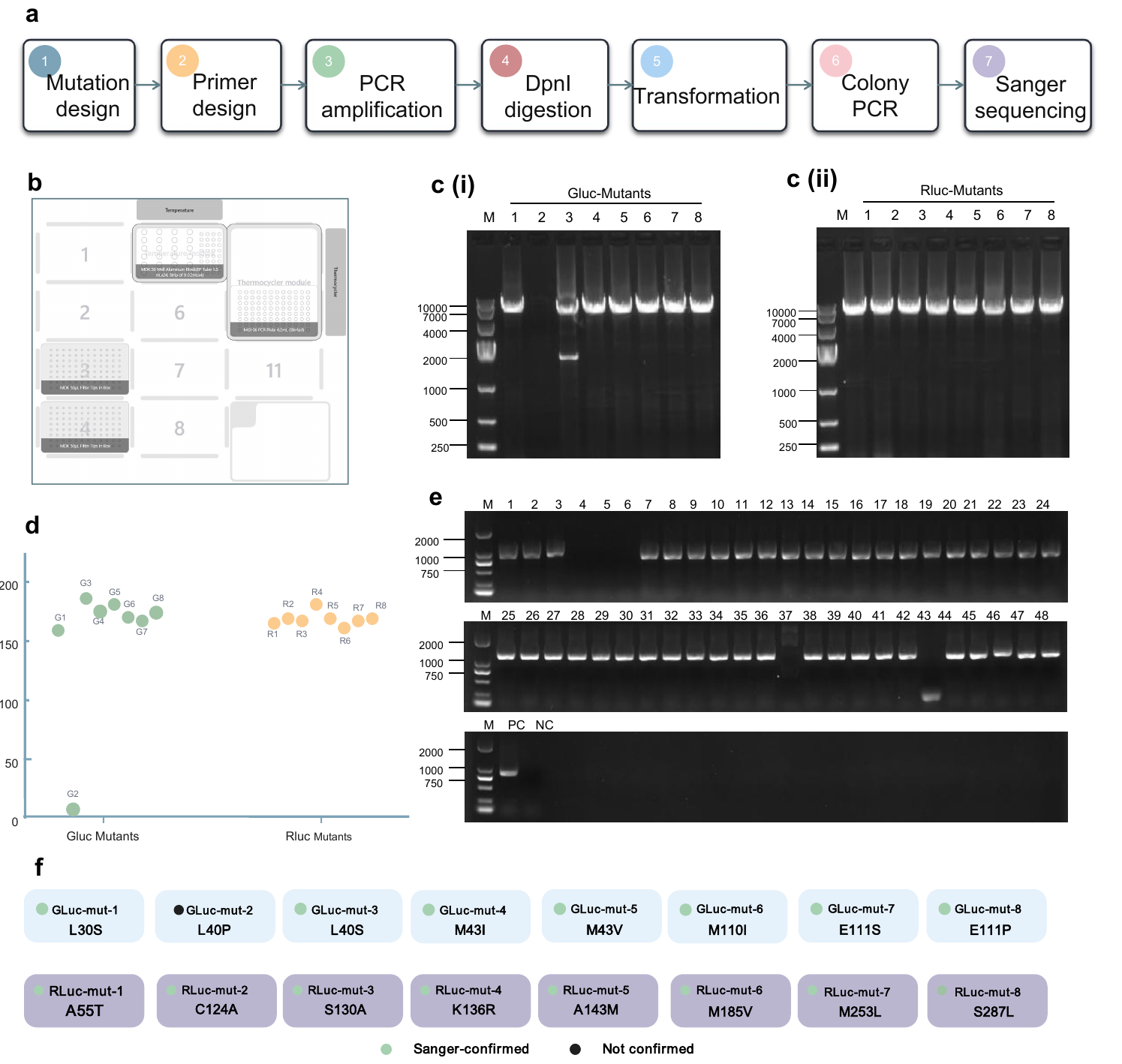}
  \caption{%
    \textbf{ProtoPilot supports parallel construction of GLuc and RLuc point mutants.}
    \textbf{(a)}~Workflow diagram for whole-plasmid mutagenesis of GLuc and RLuc.
    \textbf{(b)}~Automation deck layout for parallel multi-mutant construction.
    \textbf{(c)}~Agarose gel electrophoresis of whole-plasmid PCR products for GLuc mutants (i) and RLuc mutants (ii); M2 denotes DL10000 DNA marker. Lanes 1--8 indicate the eight designed mutants in each series: GLuc lanes 1--8 correspond to L30S, L40P, L40S, M43I, M43V, M110I, E111S and E111P, respectively, and RLuc lanes 1--8 correspond to A55T, C124A, S130A, K136R, A143M, M185V, M253L and S287L, respectively.
    \textbf{(d)}~DNA concentrations of whole-plasmid PCR products; green points indicate GLuc mutants, orange points indicate RLuc mutants, and low-yield points indicate failed or insufficient amplification.
    \textbf{(e)}~Colony PCR screening of candidate clones after transformation; M1 denotes DL2000 DNA marker, PC denotes positive control, NC denotes negative control, lanes 1--24 denote GLuc candidate clones and lanes 25--48 denote RLuc candidate clones, with three candidate clones tested per mutant. GLuc lanes 1--3, 4--6, 7--9, 10--12, 13--15, 16--18, 19--21 and 22--24 correspond to L30S, L40P, L40S, M43I, M43V, M110I, E111S and E111P, respectively; RLuc lanes 25--27, 28--30, 31--33, 34--36, 37--39, 40--42, 43--45 and 46--48 correspond to A55T, C124A, S130A, K136R, A143M, M185V, M253L and S287L, respectively.
    \textbf{(f)}~Sanger sequencing summary of designed GLuc and RLuc mutants; the central circles indicate the number of confirmed mutants in each series, green markers indicate Sanger-confirmed constructs, and the black marker indicates an unconfirmed construct.
  }
  \label{fig:fig5}
\end{figure}

Building on the wild-type plasmid construction results, we extended the evaluation to whole-plasmid mutagenesis of GLuc and RLuc to assess ProtoPilot in a parallel multi-variant construction setting. Compared with wild-type construction, site-directed mutagenesis additionally requires mutation design, primer design, and parallel management of multiple target amino-acid substitutions, placing greater demands on workflow organization and sample tracking (Fig.~\ref{fig:fig5}a).

Upon receiving the user's request to construct multiple GLuc and RLuc point mutants in parallel, ProtoPilot reasoned over the mutation targets and experimental route, then used multi-turn interaction to clarify the target sites, mutation types, screening requirements, and available laboratory reagents, consumables, and equipment. After the experimental plan was confirmed, ProtoPilot generated an executable automation workflow supporting parallel construction of multiple point mutants, with a representative deck configuration illustrating how the mutant reactions were organized into an executable physical setup (Fig.~\ref{fig:fig5}b). In total, 16 point mutants were designed: eight GLuc mutants (L30S, L40P, L40S, M43I, M43V, M110I, E111S, and E111P) and eight RLuc mutants (A55T, C124A, S130A, K136R, A143M, M185V, M253L, and S287L).

Agarose gel electrophoresis showed that all mutants except GLuc-L40P produced clear single whole-plasmid PCR bands at the expected sizes, approximately 6,048\,bp for the GLuc series and 6,423\,bp for the RLuc series (Fig.~\ref{fig:fig5}c). Quantification confirmed that the failed GLuc-L40P amplification yielded only 4.8\,ng/\textmu L, whereas the remaining 15 mutants yielded 160--187\,ng/\textmu L. The GLuc mutant concentrations for L30S, L40S, M43I, M43V, M110I, E111S, and E111P were 160, 187, 176, 182, 171, 168, and 175\,ng/\textmu L, respectively; the RLuc mutant concentrations for A55T, C124A, S130A, K136R, A143M, M185V, M253L, and S287L were 166, 170, 168, 182, 170, 162, 168, and 170\,ng/\textmu L, respectively (Fig.~\ref{fig:fig5}d).

The successfully amplified mutants were subjected to DpnI digestion, transformation, and colony PCR screening. Agarose gel electrophoresis confirmed that most candidate clones produced clear bands of the expected size suitable for downstream Sanger confirmation (Fig.~\ref{fig:fig5}e). Sanger sequencing confirmed 7 of the 8 designed GLuc mutants, with GLuc-L40P remaining unconfirmed, and all 8 designed RLuc mutants (Fig.~\ref{fig:fig5}f). These results demonstrate that ProtoPilot can support parallel construction of multiple point mutants across a complete variant-engineering workflow spanning mutation design, whole-plasmid amplification, template removal, transformation screening, and sequence confirmation.

\subsubsection{ProtoPilot refines experimental protocols iteratively through closed-loop feedback during PCA-based DNA assembly}

\begin{figure}[htbp]
  \centering
  \includegraphics[width=0.95\textwidth]{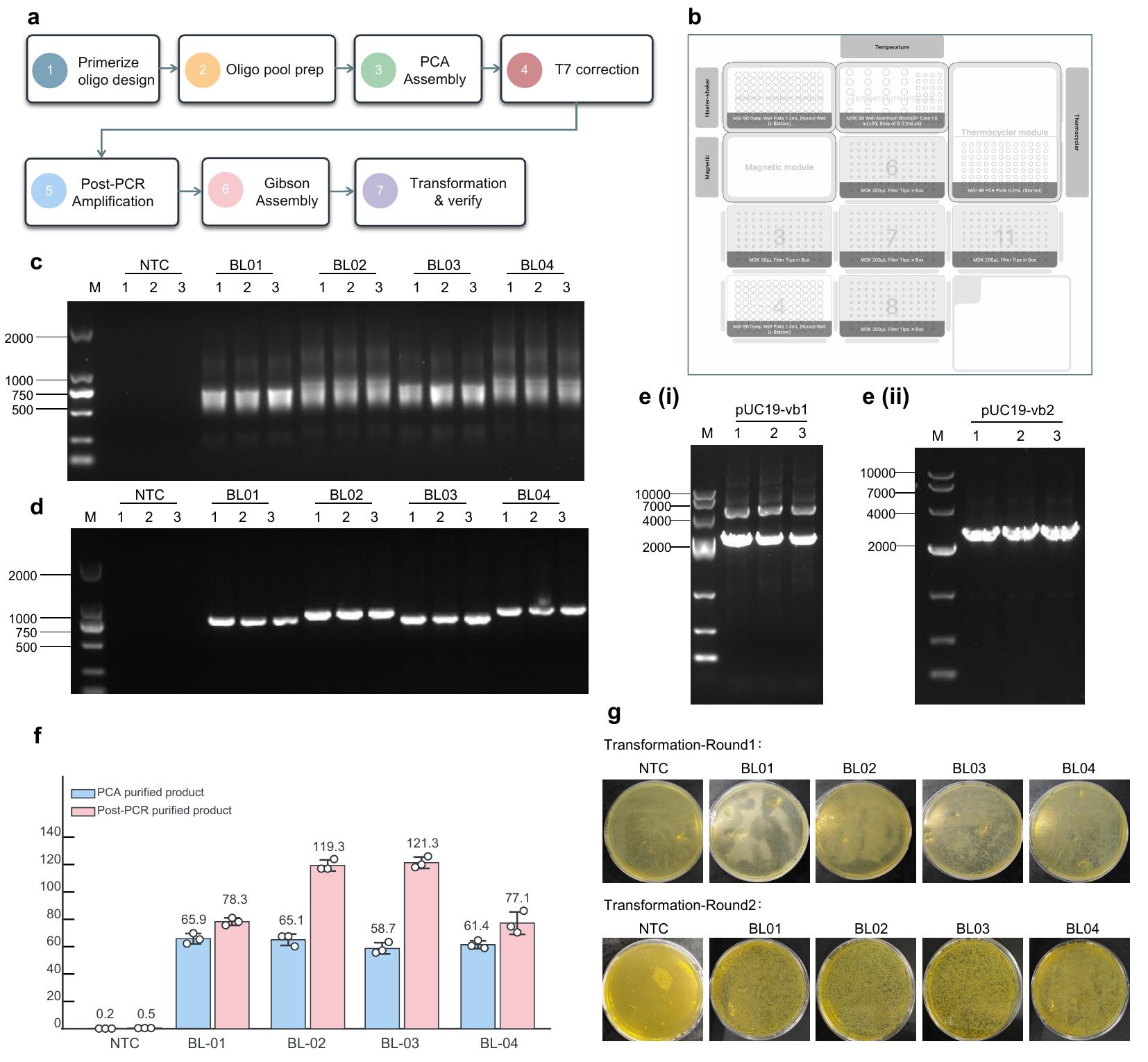}
  \caption{%
    \textbf{ProtoPilot supports parallel construction of GLuc and RLuc point mutants.}
    \textbf{(a)}~ProtoPilot supports PCA-based DNA assembly and iterative protocol refinement through feedback.
    \textbf{(b)}~Automation deck layout for the PCA DNA assembly experiment.  
    \textbf{(c)}~Agarose gel electrophoresis of PCA assembly products for target fragments BL01-BL04; M1 denotes DL2000 DNA marker, NTC denotes no-template control, and lanes 1-3 denote three independent replicates for each fragment. 
    \textbf{(d)}~Agarose gel electrophoresis of post-PCR products after T7 Endonuclease I correction; M1 denotes DL2000 DNA marker, NTC denotes no-template control, and lanes 1-3 denote three independent replicates for each fragment. 
    \textbf{(e)}~Optimization of pUC19 vector-backbone PCR linearization; M2 denotes DL10000 DNA marker, and lanes 1-3 denote three independent replicates. The first preparation, pUC19-vb1, showed a prominent nonspecific amplification band in addition to the target band (i), whereas primer redesign yielded pUC19-vb2 with a clear single target band (ii). 
    \textbf{(f)}~DNA concentrations of purified PCA and post-PCR products; blue bars indicate purified PCA products, pink bars indicate purified post-PCR products, points indicate independent replicates, and values above bars indicate mean concentrations.
    \textbf{(g)}~Plate images before and after revision of the transformation protocol; NTC denotes negative control, and BL01-BL04 denote assembly products for the four target fragments. In the first transformation round, both NTC and sample plates showed abnormal growth; after protocol revision, the NTC plate showed no colony growth and the BL01-BL04 plates produced single colonies suitable for downstream screening.
  }
  \label{fig:fig6}
\end{figure}

Beyond plasmid construction and site-directed mutagenesis, we next evaluated polymerase cycling assembly (PCA)-based DNA assembly as a more complex long-form construction workflow. Starting from Primerize-designed oligonucleotide pools, this workflow involves oligo-pool preparation, PCA assembly, T7 Endonuclease I correction, post-PCR amplification, Gibson assembly, transformation, and clone validation (Fig.~\ref{fig:fig6}a). The PCA workflow contains more intermediate products, a longer chain of experimental steps, and more complex sample relationships than conventional PCR or single-plasmid construction, making it a demanding test case for multi-module DNA assembly.

Upon receiving the user's request for parallel assembly and cloning of multiple target fragments, ProtoPilot reasoned over the assembly strategy and validation route, then used multi-turn interaction to clarify the target sequences, assembly mode, validation checkpoints, and available laboratory reagents, consumables, and equipment. ProtoPilot selected four target fragments of similar length but different origins for parallel assembly: BL01, an EGFP coding sequence of 698\,bp, and three regions from A501\_Ec.CPS, namely BL02, BL03, and BL04, with lengths of 802, 713, and 800\,bp, respectively. The corresponding automation layout shows a representative deck configuration illustrating how the long DNA assembly workflow was decomposed and mapped into an executable automation environment (Fig.~\ref{fig:fig6}b). Each fragment was assembled in three independent replicates to evaluate workflow robustness.

PCA assembly products for all four target fragments showed bands at the expected sizes, whereas the no-template control showed no obvious amplification (Fig.~\ref{fig:fig6}c). After T7 Endonuclease I correction, post-PCR amplification produced clear bands of the expected sizes for all four fragments, confirming that the assembled and corrected products were suitable for downstream cloning (Fig.~\ref{fig:fig6}d). Quantification after purification ($n = 3$ replicates per group) showed PCA product concentrations of 65.93~$\pm$~3.83, 65.07~$\pm$~4.22, 58.73~$\pm$~4.12, and 61.40~$\pm$~2.88\,ng/\textmu L for BL01--BL04, respectively, and post-PCR product concentrations of 78.27~$\pm$~2.70, 119.33~$\pm$~4.04, 121.33~$\pm$~4.16, and 77.13~$\pm$~8.15\,ng/\textmu L for BL01--BL04, respectively. No-template controls remained near background (approximately 0.2\,ng/\textmu L for PCA and 0.5\,ng/\textmu L for post-PCR), indicating no detectable contamination during the workflow (Fig.~\ref{fig:fig6}f).

Vector preparation and downstream assembly revealed two iterative optimization cycles. For vector preparation, pUC19-vb1 produced the expected 2,619\,bp target band but also showed a prominent nonspecific band in the approximately 4,000--7,000\,bp region (Fig.~\ref{fig:fig6}e(i)); ProtoPilot analyzed the failure mode and redesigned the primers to improve specificity, and the second preparation, pUC19-vb2, produced a clear single band at 2,619\,bp (Fig.~\ref{fig:fig6}e(ii)). The four post-PCR fragments were then assembled with pUC19-vb2 by Gibson assembly and transformed. The first transformation showed a clear failure: colonies appeared on the negative-control plate, and the BL01--BL04 sample plates exhibited extensive or lawn-like growth that precluded single-colony picking (Fig.~\ref{fig:fig6}g). ProtoPilot attributed this to ampicillin inactivation and excessive plating volume, regenerated a revised protocol using freshly prepared ampicillin plates and reduced plating volume, and re-executed the transformation. The second round produced no negative-control colonies and yielded distinguishable single colonies across all four target plates (Fig.~\ref{fig:fig6}g), demonstrating ProtoPilot's capacity for failure diagnosis, protocol reconstruction, and iterative re-execution in response to real wet-lab abnormalities. Complete replicate plate results from the first and second transformation rounds are shown in Extended Data Fig. 17.

Colony PCR screening across three transformation replicate plates, with eight colonies picked per plate, showed that 93 of 96 candidate clones were PCR-positive, corresponding to an overall positivity rate of 96.9\% (
). Sanger sequencing of four positive clones per replicate plate confirmed 8, 6, 9, and 6 sequence-correct clones for BL01, BL02, BL03, and BL04, respectively (
). All four target DNA sequences produced at least one Sanger-confirmed correct clone in an independent replicate, demonstrating successful construction of each target sequence.

Together, these results show that ProtoPilot can support a long-form DNA construction workflow spanning oligonucleotide-pool design, fragment assembly, correction and amplification, vector preparation, homologous recombination, and clone screening. Critically, ProtoPilot identified and resolved two distinct experimental failures during execution, confirming that the system is not limited to one-shot protocol generation but can perform multi-module execution, failure handling, and iterative optimization in complex synthetic-biology construction workflows.

\section{Discussion}

Here we present ProtoPilot, a self-evolving multi-agent system for automating life-science experimental workflows from natural-language intent to validated SOPs, instrument-executable code, physical execution and feedback-guided refinement. Rather than treating protocol writing, code generation and robotic execution as separate tasks, ProtoPilot formulates laboratory automation as a stateful workflow in which biological context, operational constraints, validation outcomes and experimental feedback are maintained across the entire process. Our results show that this design enables accurate protocol generation, reliable protocol-to-code translation and practical wet-lab execution across diverse experimental settings.

The core technical contribution of ProtoPilot is its integration of hierarchical agent orchestration with reusable operational skills. The novel multi-agent collaboration mechanism allows the system to decompose long SOPs into manageable modules while maintaining global workflow state, sample lineage, parameter dependencies and quality-control logic. This is essential for real experimental SOPs, where correctness depends not only on local step quality but also on consistency across distant procedural sections. In parallel, the self-evolving skill library used by the Coding Agent encodes expert operational knowledge that is often absent from protocol texts or SDK documentation. By organizing this knowledge into parameter-confirmation, deck-layout planning and code-specification skills, ProtoPilot bridges the gap between a biologically valid SOP and a physically executable robotic program.

To rigorously evaluate such a system, we constructed an expert-grounded benchmark and a comprehensive evaluation framework for autonomous wet-lab experimentation. The benchmark combines diverse synthetic-biology and molecular-biology scenarios with gold-standard protocol data, wet-lab expert experience, task-specific metrics, rubric-based validators, device-level validity gates and real experimental tests. Rather than treating protocol generation as an isolated text task, the framework assesses whether generated workflows satisfy biological intent, preserve methodological consistency, instantiate sound parameters and remain procedurally complete. It further tests whether those workflows can be translated into SDK-exportable, instrument-compatible command objects, and whether they produce interpretable biological readouts that can support revision after failed or suboptimal steps. In this way, the benchmark translates expert wet-lab judgment into measurable requirements for experimental automation, and underscores the gap between the capabilities of existing general‑purpose or specialized AI agents and those required for truly autonomous wet‑lab experimentation.

Across 294 real-world synthetic biology tasks, ProtoPilot achieved superior protocol quality across different complexity levels. This improvement was corroborated by blind expert evaluation, suggesting that the generated SOPs were not merely more complete according to automatic metrics but also more consistent with expert expectations for experimental feasibility, procedural clarity and biological validity. These results support the importance of maintaining an explicit workflow state and validating each generated module before it is integrated into the final SOP. In particular, the hierarchical generation strategy reduced the tendency of LLM agents to produce locally plausible but globally inconsistent protocols.

ProtoPilot also generated safe, executable instrument code with robust generalizability across tested device settings. A central difficulty in laboratory code generation is that syntactic correctness alone is insufficient: a script may compile while still violating the validated protocol, misaligning sample or well mappings, mishandling tips, placing labware in inaccessible positions or executing module states in an unsafe order. ProtoPilot addresses these failure modes by combining skill-mediated code synthesis with rubric-based validation of protocol--code alignment, deck and well-map consistency, liquid-handling completeness, SDK compliance and device-state constraints. The cross-device performance indicates that encoding stable expert rules, rather than relying only on protocol-specific prompts or SDK surface syntax, provides a transferable basis for adapting automation to heterogeneous instruments and software interfaces.

The wet-lab experiments provided a more stringent assessment of ProtoPilot than in silico evaluation alone. Across experiments of increasing complexity, ProtoPilot generated and executed workflows that were reliable in practice, demonstrating that its outputs can satisfy real laboratory constraints rather than only text-based benchmarks. The luciferase plasmid construction and mutant-engineering studies further show that ProtoPilot can support multi-step molecular biology workflows in which correct execution requires coordination among reagent preparation, DNA manipulation, purification, transformation or downstream validation steps. These experiments suggest that ProtoPilot can contribute to the build stage of synthetic biology workflows, where protocol detail, execution fidelity and reproducibility are all critical.

A further distinguishing feature of ProtoPilot is its ability to refine protocols from empirical feedback. In the PCA-based DNA assembly study, experimental outcomes were returned to the system and used to guide subsequent protocol modification. This closed-loop behavior is important because wet-lab automation rarely succeeds as a one-shot generation problem: reaction efficiency, assembly quality, sample loss and execution failures often require iterative adjustment. By updating the workflow state and invoking the appropriate expert agents after feedback is received, ProtoPilot provides a practical mechanism for adaptive protocol optimization rather than treating each experiment as an isolated prompt-response interaction.

Several limitations remain. First, although the current benchmark covers 294 real-world synthetic biology tasks, its scale and diversity can be further expanded. Future benchmarks should include broader assay families, additional laboratory instruments, more heterogeneous labware configurations, different biological systems and more challenging failure modes. Second, ProtoPilot currently focuses primarily on wet-lab automation after a user-defined experimental objective has been provided. A natural next step is to integrate dry-lab modules for hypothesis generation, sequence or construct design, computational modeling, data analysis and optimization, thereby establishing a more complete dry--wet closed loop. Such integration would allow ProtoPilot not only to execute and refine experiments, but also to participate in the upstream design and prioritization of experimental hypotheses.

In summary, ProtoPilot provides a framework for stateful, closed-loop automation of life-science experiments. By integrating agentic planning, long-context SOP generation, expert-skill-mediated code synthesis, validation, robotic execution and feedback-driven refinement, ProtoPilot helps bridge the gap between experimental intent and physical laboratory execution. These principles provide a foundation for future AI-native laboratories in which computational reasoning and automated experimentation are integrated into a continuous, hypothesis-driven research cycle.

\section*{Materials}
\noindent \textbf{Bacterial strains, plasmids and culture media.}

\textit{Escherichia coli} BL21(DE3) was used as the host strain for plasmid construction, site-directed mutagenesis and clone propagation, and \textit{E. coli} DH5$\alpha$ was used for de novo PCA assembly and cloning. The pET-21b(+) vector was used as the expression backbone for constructing wild-type Gaussia luciferase and Renilla luciferase plasmids, designated pET21b-GLuc-WT and pET21b-RLuc-WT, respectively. The pUC19 vector was used as the cloning backbone for PCA-based de novo gene assembly. Sequence-confirmed pET21b-GLuc-WT and pET21b-RLuc-WT plasmids were subsequently used as templates for whole-plasmid site-directed mutagenesis.
Bacteria were cultured in liquid or solid LB medium. Ampicillin was added where required at a final concentration of 100~\textmu g/mL from a 100 mg/mL stock solution. LB agar plates containing ampicillin were used for selection after transformation, and liquid LB medium containing ampicillin was used for clone propagation and overnight culture. Unless otherwise specified, bacterial cultures were incubated at 37~\textdegree C; liquid cultures were shaken at 200 rpm, and agar plates were incubated inverted for 12 h.

\noindent \textbf{Enzymes, kits and reagents.}

PCR amplification was performed using 2$\times$ Phanta Flash Master Mix for high-fidelity PCR and 2$\times$ Rapid Taq Master Mix for colony PCR. Methylated parental plasmid DNA was removed using SwiftCut DpnI, and homologous recombination assembly was performed using the ClonExpress II kit; these reagents were purchased from Vazyme (Nanjing, China). Mismatch correction in the PCA workflow used T7 Endonuclease I (Beyotime, Shanghai, China). Agarose gel electrophoresis used 4S GelRed nucleic acid stain (Sangon Biotech, Shanghai, China) and DL10000 or DL2000 DNA markers (Takara, Japan). Gel-excised DNA fragments were purified using the OMEGA E.Z.N.A. Gel Extraction Kit (OMEGA Bio-tek, USA). PCA and post-PCR products were purified using MGIEasy DNA purification magnetic beads (MGI, Shenzhen, China) and quantified by Qubit fluorometry (Thermo Fisher, USA). Chemically competent \textit{E. coli} BL21(DE3) and \textit{E. coli} DH5$\alpha$ cells were purchased from Solarbio (Beijing, China), and LB medium (Sangon Biotech, Shanghai, China) was used for recovery and culture.









\printbibliography

\section*{Acknowledgements}

This work was jointly supported by Shanghai Artificial Intelligence Laboratory and Genoria AI Technology Co., Ltd.


\section*{Competing Interests}

The authors declare no competing interests.


\section*{Appendix}
\setcounter{figure}{0}
\renewcommand{\thefigure}{\arabic{figure}}
\captionsetup[figure]{name={\textbf{Extended Data Fig.}}}

\setcounter{table}{0}
\renewcommand{\thetable}{\arabic{table}}
\captionsetup[table]{name={\textbf{Extended Data Table}}}




\begin{figure}[htbp]
  \centering
  \begin{subfigure}{\textwidth}
    \includegraphics[width=\linewidth]{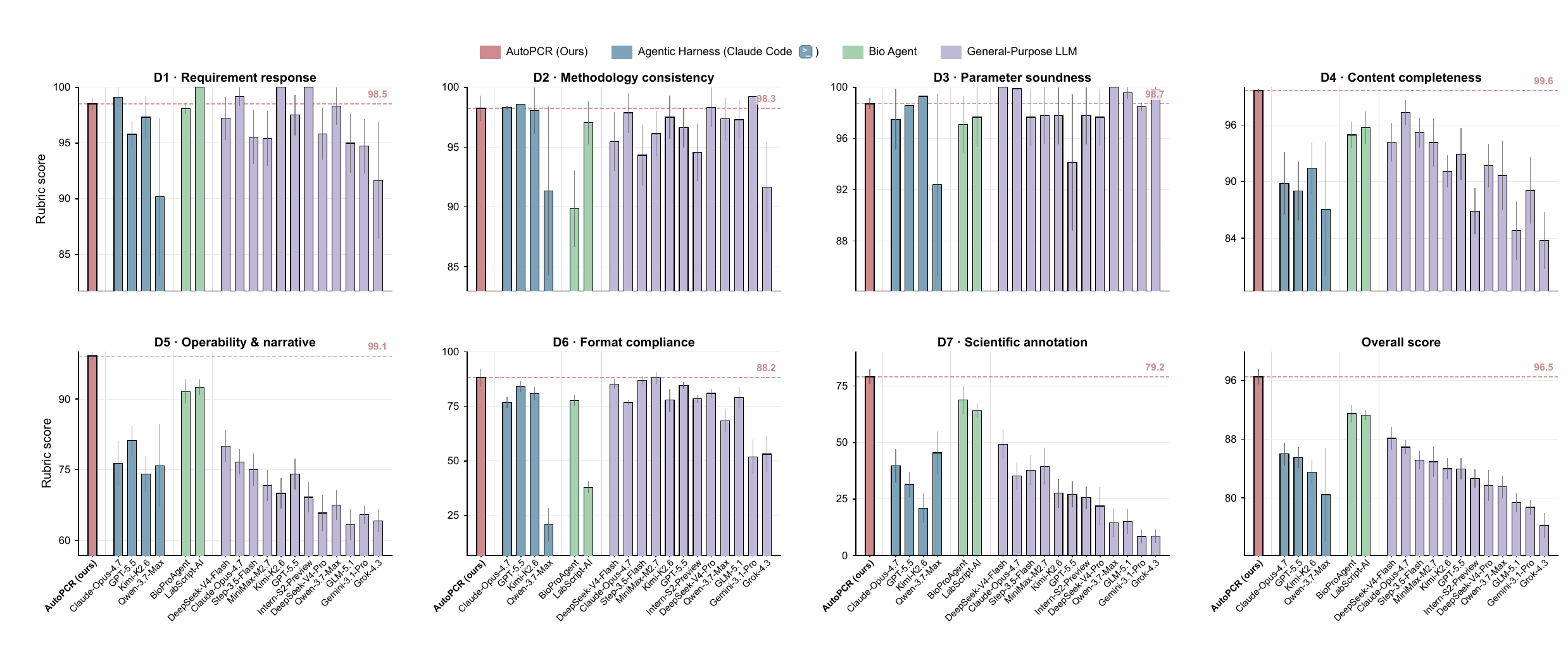}
    \caption{L1 (basic protocols, $n_{\text{ref}} = 15$)}
  \end{subfigure}
  \vspace{0.8em}
  \begin{subfigure}{\textwidth}
    \includegraphics[width=\linewidth]{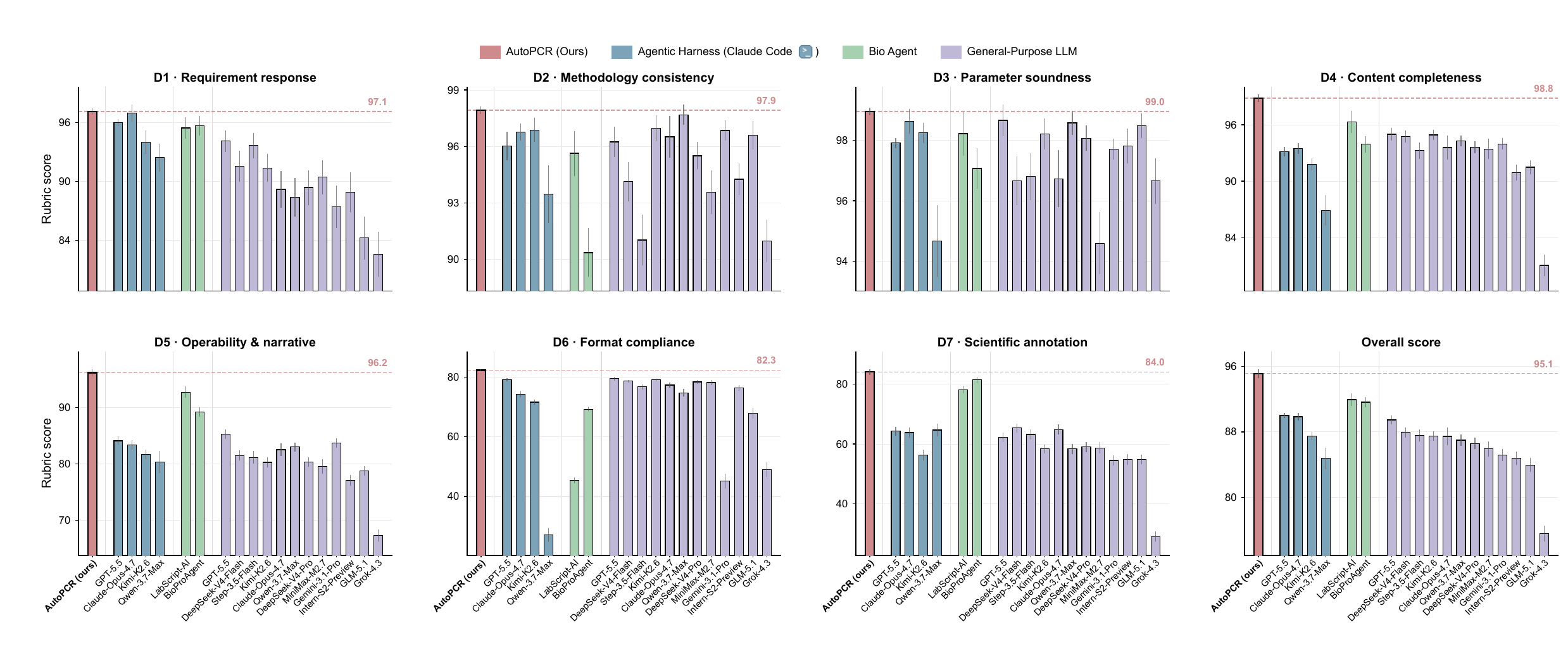}
    \caption{L2 (intermediate protocols, $n_{\text{ref}} = 165$)}
  \end{subfigure}
  \vspace{0.8em}
  \begin{subfigure}{\textwidth}
    \includegraphics[width=\linewidth]{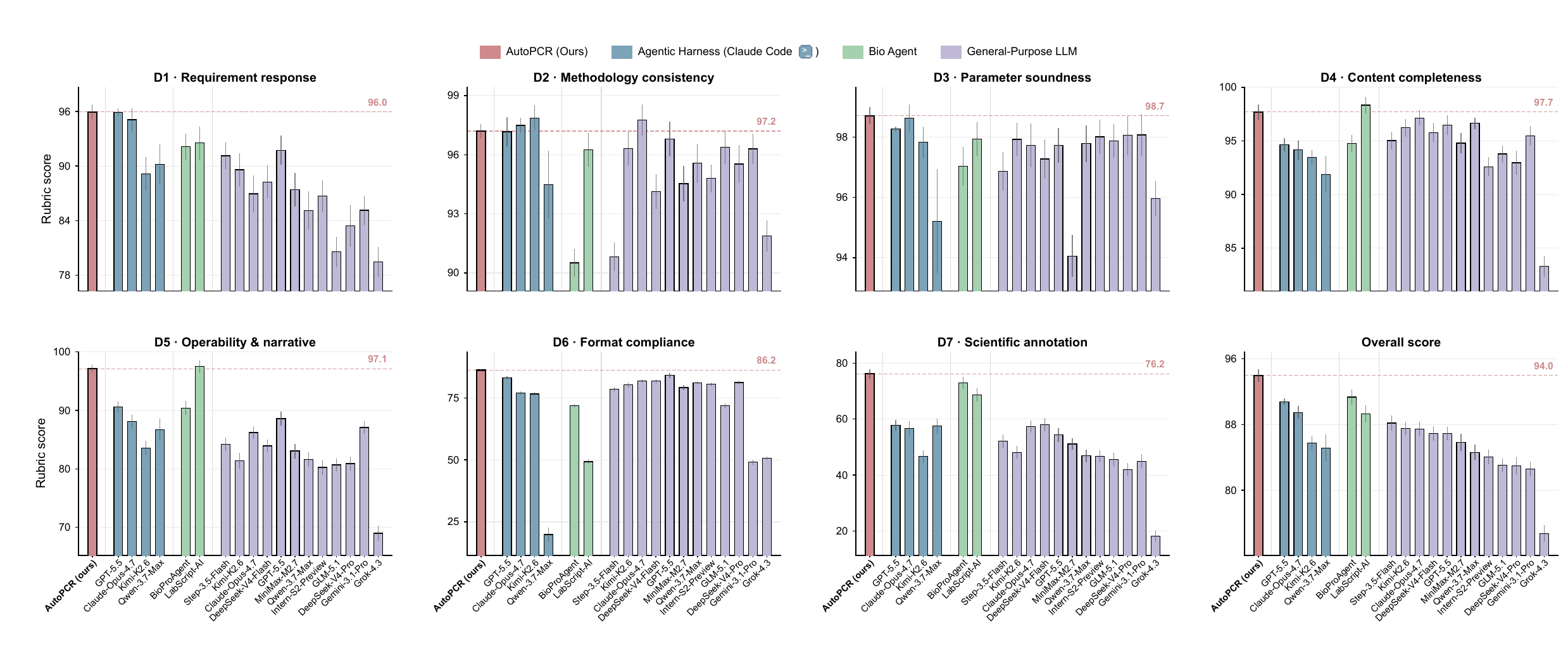}
    \caption{L3 (advanced protocols, $n_{\text{ref}} = 114$)}
  \end{subfigure}
  \caption{%
    \textbf{Per-difficulty rubric score profiles (Supplementary to Fig.~\ref{fig:fig2}a).}
    Mean rubric scores (D1--D7 and Overall; mean $\pm$ s.e.m.) broken down by
    difficulty tier L1, L2, and L3.
    ProtoPilot overall scores: L1 = 96.5, L2 = 95.1, L3 = 94.0, demonstrating
    consistent high performance across difficulty levels.
    Harness-augmented systems (Claude Code) show markedly lower D7 scores, particularly at L3 where domain-specific
    biological constraints are most demanding.
  }
  \label{fig:app_fig2a}
\end{figure}

\begin{figure}[htbp]
  \centering
  \includegraphics[width=\textwidth]{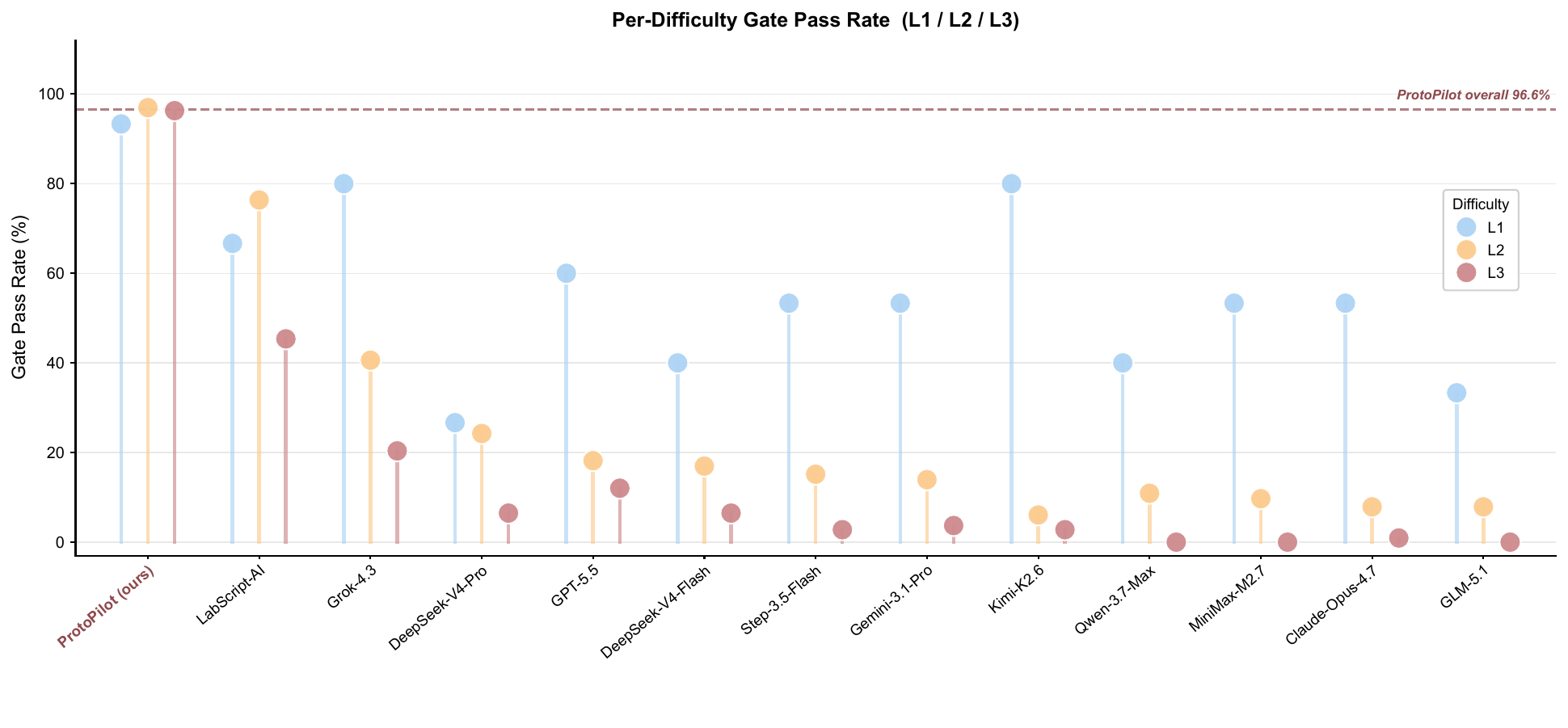}
  \caption{%
    \textbf{Per-difficulty gate pass rate across 13 systems
    (Supplementary to Fig.~\ref{fig:fig2}b).}
    Gate pass rate (\%) for each system stratified by difficulty tier
    (L1 = blue, L2 = amber, L3 = red).
    Each dot represents the pass rate for one system--difficulty combination.
    ProtoPilot gate rates: L1 = 93.3\%, L2 = 97.0\%, L3 = 96.3\%.
    The dashed horizontal line indicates ProtoPilot's overall gate rate (96.6\%).
    General-purpose LLMs reach up to 80.0\% on L1 but fall below 41\% on L2
    and below 21\% on L3.
  }
  \label{fig:app_b1}
\end{figure}

\begin{figure}[htbp]
  \centering
  \includegraphics[width=0.85\textwidth]{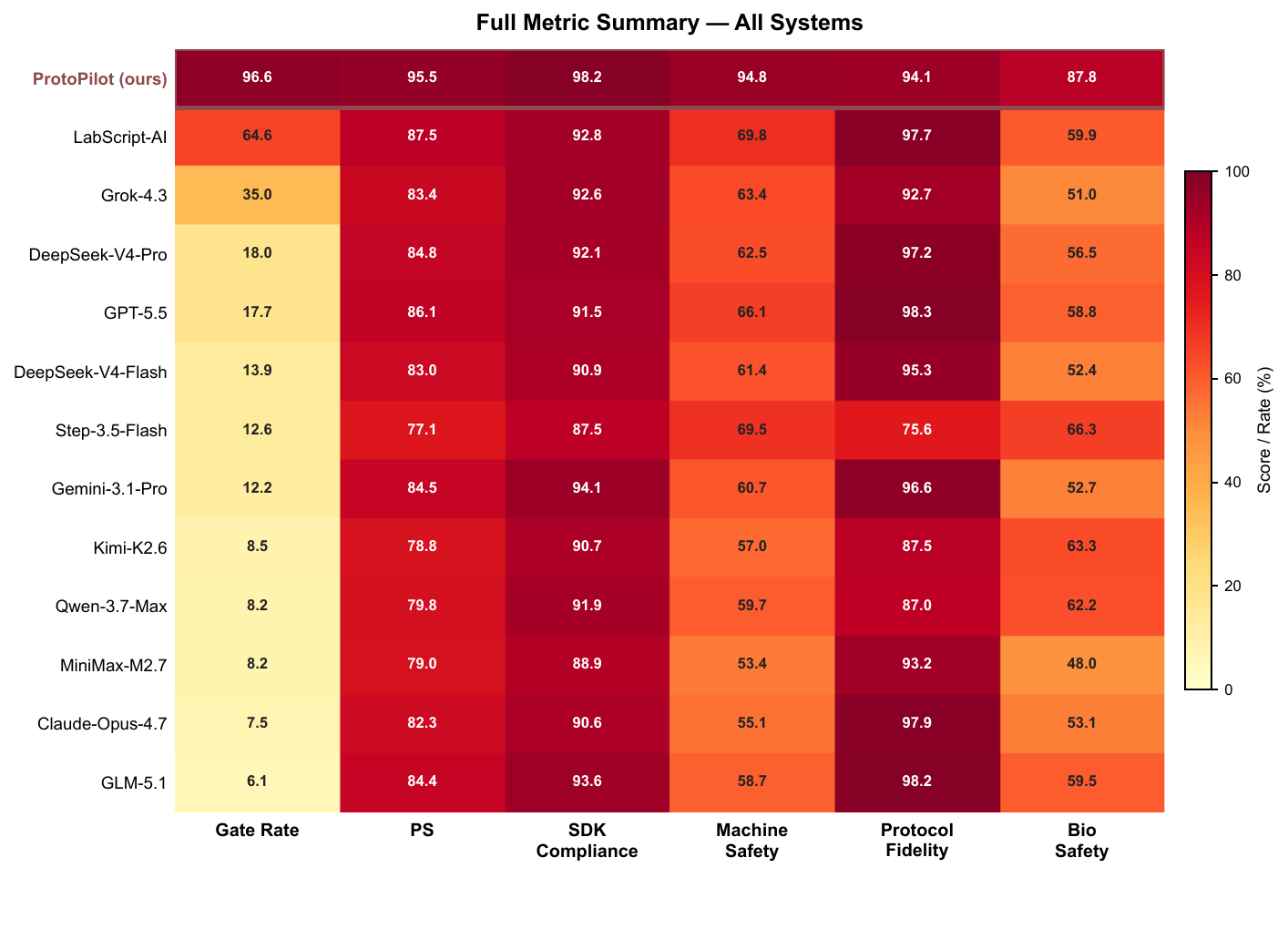}
  \caption{%
    \textbf{Full-metric summary heatmap for all 13 systems
    (Supplementary to Fig.~\ref{fig:fig2}b).}
    Colour-encoded performance matrix across six evaluation
    metrics: Gate Rate, Code Score, SDK Compliance, Machine Safety,
    Protocol Fidelity, and Bio Safety.
    ProtoPilot leads on five of six metrics.
  }
  \label{fig:app_b2}
\end{figure}

\begin{figure}[htbp]
  \centering
  \includegraphics[width=\textwidth]{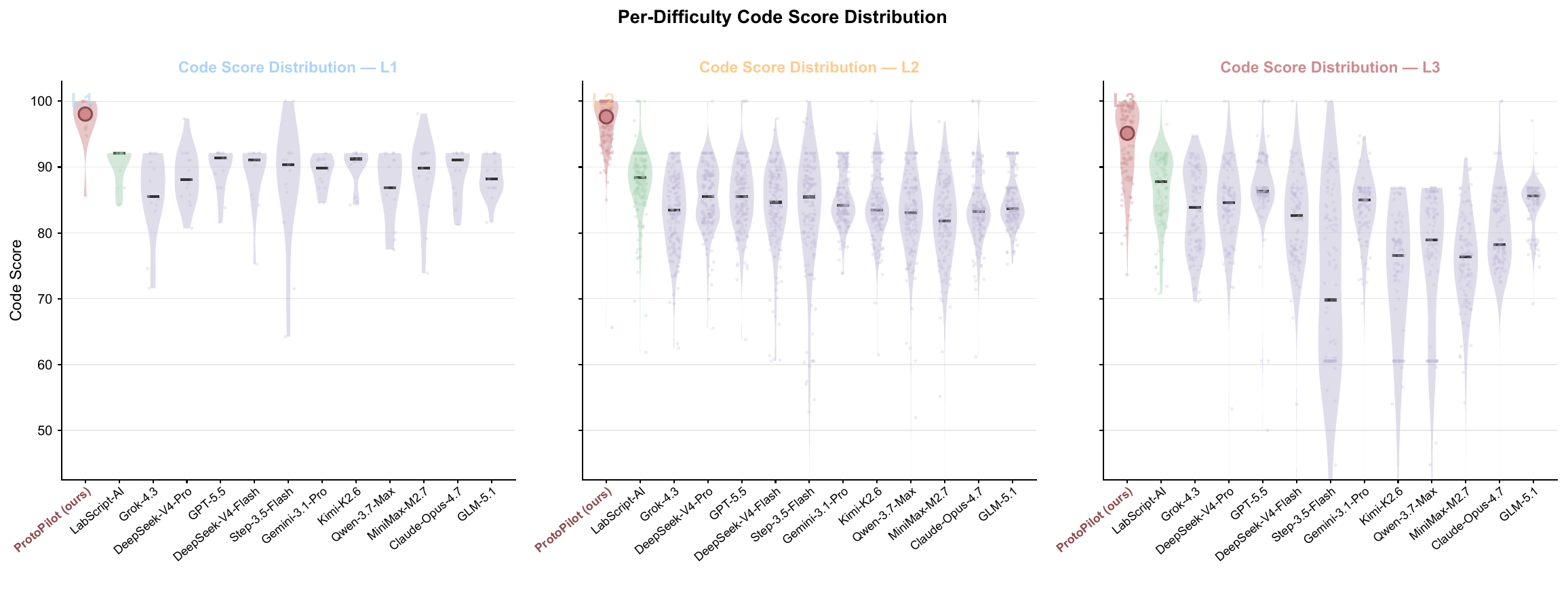}
  \caption{%
    \textbf{Per-difficulty Code Score distributions
    (Supplementary to Fig.~\ref{fig:fig2}b).}
    Violin plots with overlaid jitter showing the per-case Code Score
    distribution for each system, stratified by difficulty tier
    (L1, L2, L3 from left to right).
    Violin fill colour encodes system category (red = ProtoPilot;
    green = bio agent;
    lavender = general-purpose LLMs).
    The filled circle with dark-red edge marks the ProtoPilot median.
    ProtoPilot maintains a narrow, high-scoring distribution at all three
    difficulty levels, whereas general-purpose LLMs exhibit wider variance
    and lower medians at L3.
  }
  \label{fig:app_b3}
\end{figure}

\begin{figure}[htbp]
  \centering
  \includegraphics[width=\textwidth]{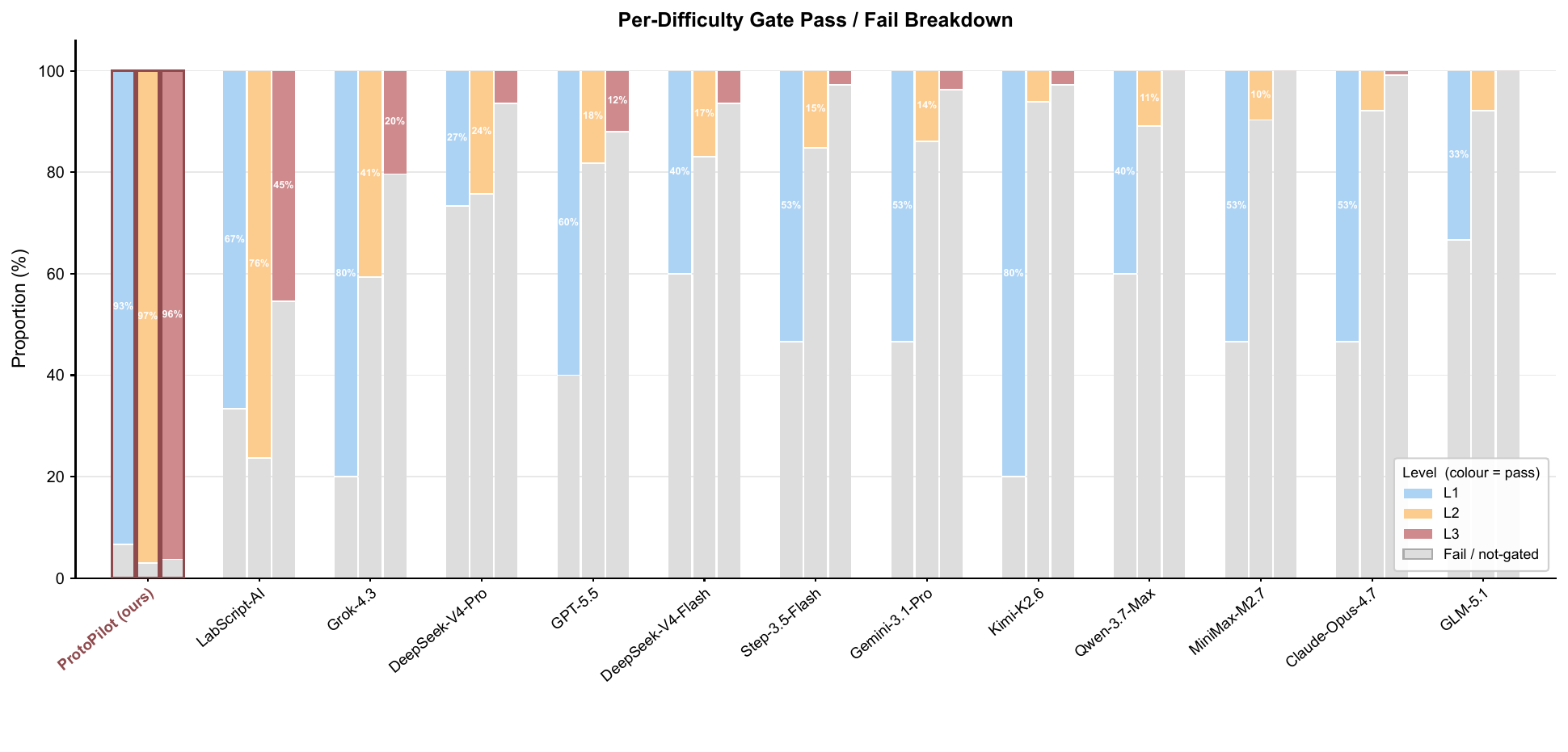}
  \caption{%
    \textbf{Per-difficulty gate pass/fail breakdown across 13 systems
    (Supplementary to Fig.~\ref{fig:fig2}b).}
    Stacked bar chart showing the proportion of gate-passing (coloured) and
    gate-failing (grey) executions per system per
    difficulty tier (L1 = blue, L2 = amber, L3 = red).
    Percentage labels inside bars indicate pass proportions.
    ProtoPilot bars are outlined in dark red.
  }
  \label{fig:app_b4}
\end{figure}

\begin{figure}[htbp]
  \centering
  \includegraphics[width=\textwidth]{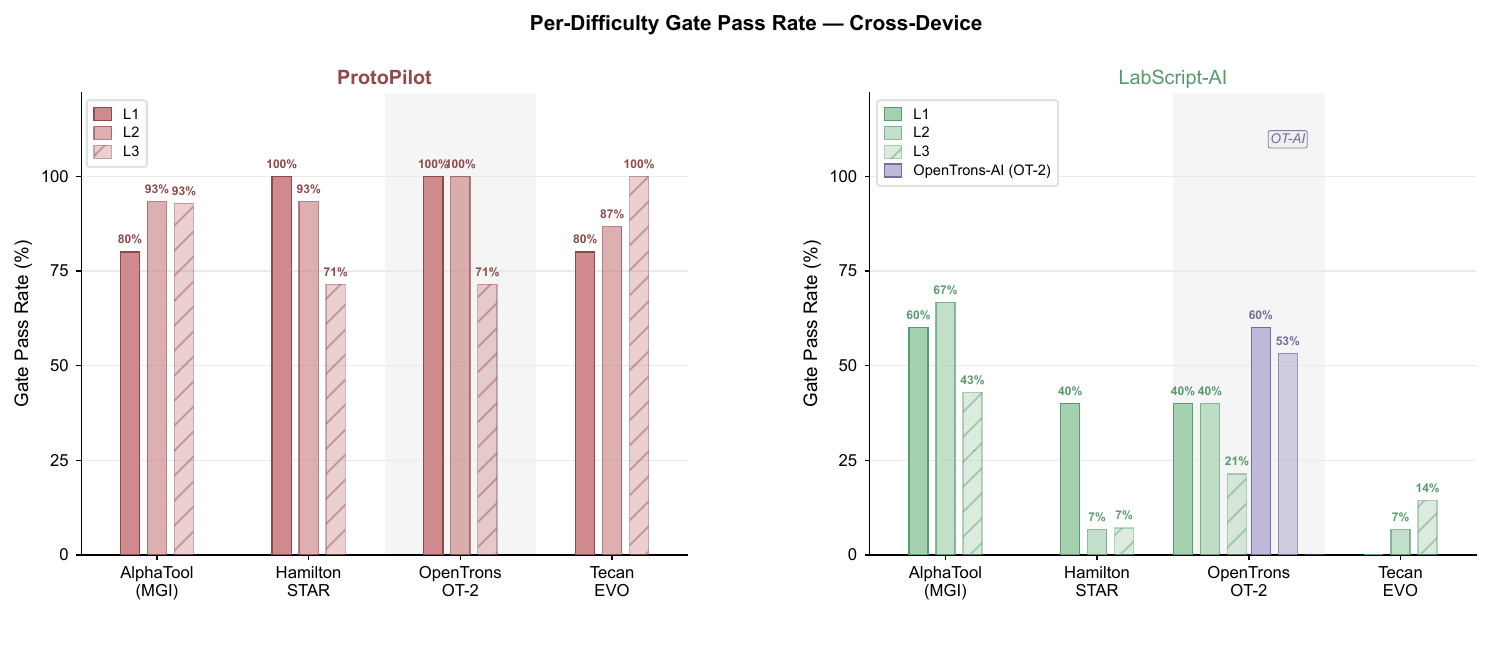}
  \caption{%
    \textbf{Per-difficulty gate pass rate across four automation platforms
    (Supplementary to Fig.~\ref{fig:fig2}c).}
    Grouped bar charts showing gate pass rate (\%) for ProtoPilot (left) and
    LabScript-AI (right) stratified by difficulty tier (L1, full opacity;
    L2, intermediate; L3, hatched) across AlphaTool (MGI), Hamilton STAR,
    OpenTrons OT-2, and Tecan EVO ($n = 5$, $15$, and $14$ tasks at L1, L2,
    and L3, respectively).
  }
  \label{fig:app_c1}
\end{figure}

\begin{figure}[htbp]
  \centering
  \includegraphics[width=\textwidth]{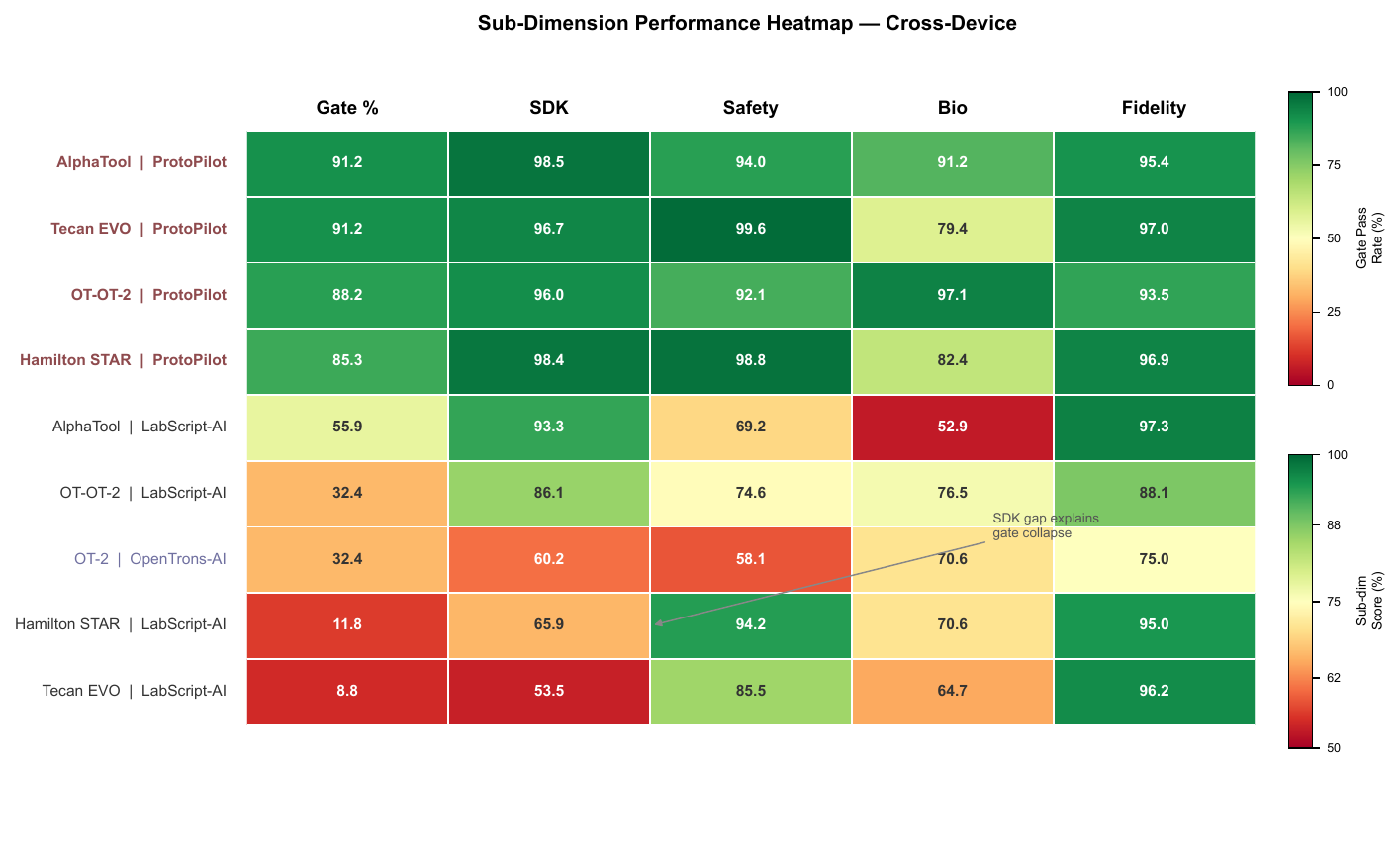}
  \caption{%
    \textbf{Sub-dimension performance heatmap across four automation platforms
    (Supplementary to Fig.~\ref{fig:fig2}c).}
    Colour-encoded matrix of Gate Pass Rate, SDK Compliance, Machine Safety,
    Bio Safety, and Protocol Fidelity for ProtoPilot and LabScript-AI on each
    of the four devices, with OpenTrons-AI included for OT-2.
    Rows are sorted by Gate Pass Rate in descending order; ProtoPilot rows are
    bold.
    Numerical values are annotated in each cell.
  }
  \label{fig:app_c2}
\end{figure}

\begin{figure}[htbp]
  \centering
  \includegraphics[width=\textwidth]{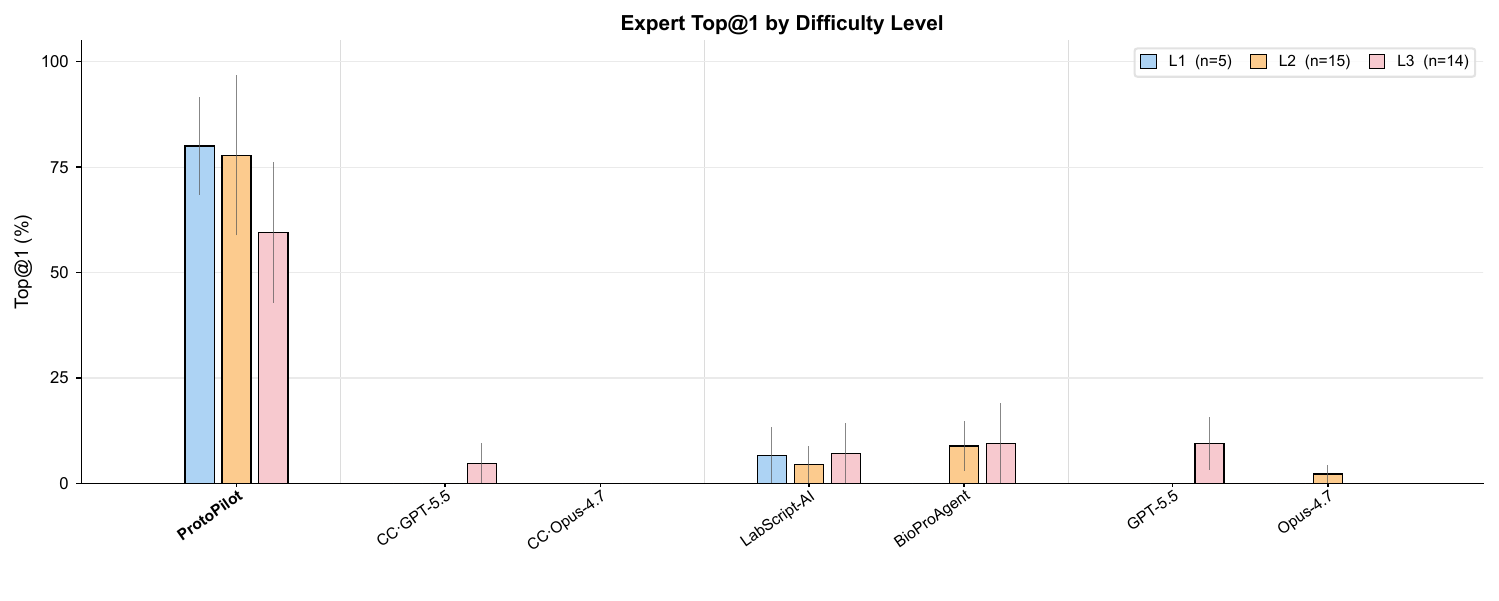}
  \caption{%
    \textbf{Per-difficulty expert top-1 preference rate
    (Related to Fig.~\ref{fig:fig2}d).}
    Top@1 rate (\%) for each system stratified by difficulty tier.
    Three bars per system correspond to L1 ($n = 5$ queries), L2 ($n = 15$ queries),
    and L3 ($n = 14$ queries).
    ProtoPilot top-1 rate by difficulty tier (mean top-1 rate across raters):
    L1 = 80.0\%, L2 = 77.8\%, L3 = 59.5\%.
    All other systems fall below 12\% at every tier.
    Error bars indicate s.e.m. across the three expert raters.
  }
  \label{fig:app_d1}
\end{figure}

\begin{figure}[htbp]
  \centering
  \includegraphics[width=\textwidth]{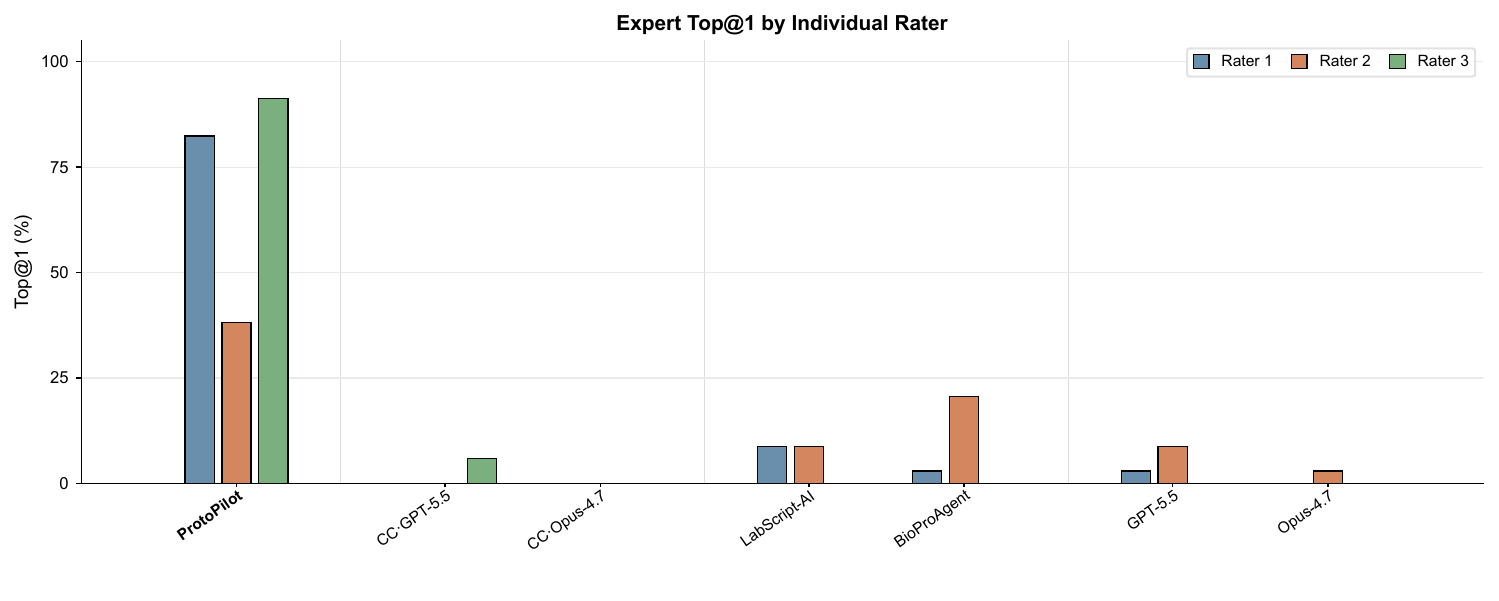}
  \caption{%
    \textbf{Per-rater expert top-1 preference rate
    (Supplementary to Fig.~\ref{fig:fig2}d).}
    Top@1 rate (\%) for each system broken down by individual rater
    (Rater~1, Rater~2, Rater~3) across all 34 queries.
    ProtoPilot Top@1 per rater: 82.4\%, 38.2\%, 91.2\%
    (mean 70.6\%); the inter-rater spread reflects differences in
    individual stringency thresholds rather than systematic disagreement,
    as all other systems remain below 21\% for every rater.
  }
  \label{fig:app_d2}
\end{figure}

\begin{figure}[htbp]
  \centering
  \includegraphics[width=0.75\textwidth]{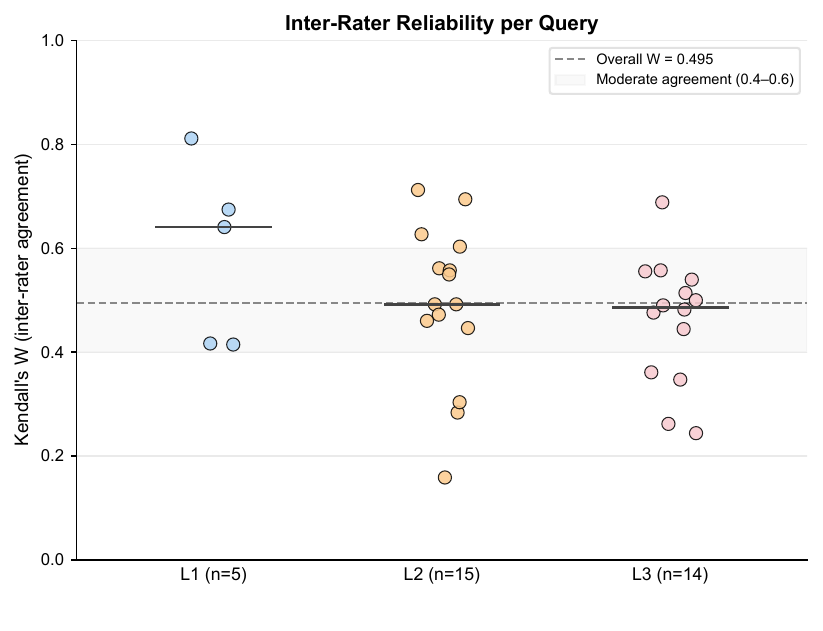}
  \caption{%
    \textbf{Inter-rater agreement per query
    (Supplementary to Fig.~\ref{fig:fig2}d).}
    Kendall's $W$ coefficient of concordance computed per query across the three
    expert raters ($n_{\text{experts}} = 3$, $n_{\text{queries}} = 34$,
    $n_{\text{systems}} = 7$).
    Each dot represents one query; horizontal bars indicate the per-level median.
    Overall Kendall's $W = 0.495$, indicating moderate inter-rater agreement.
    Levels: L1 ($n = 5$ queries), L2 ($n = 15$), L3 ($n = 14$).
    Reference band (grey shading) indicates the range of moderate agreement (0.4--0.6).
  }
  \label{fig:app_d3}
\end{figure}

\begin{figure}[htbp]
  \centering
  \includegraphics[width=0.78\textwidth]{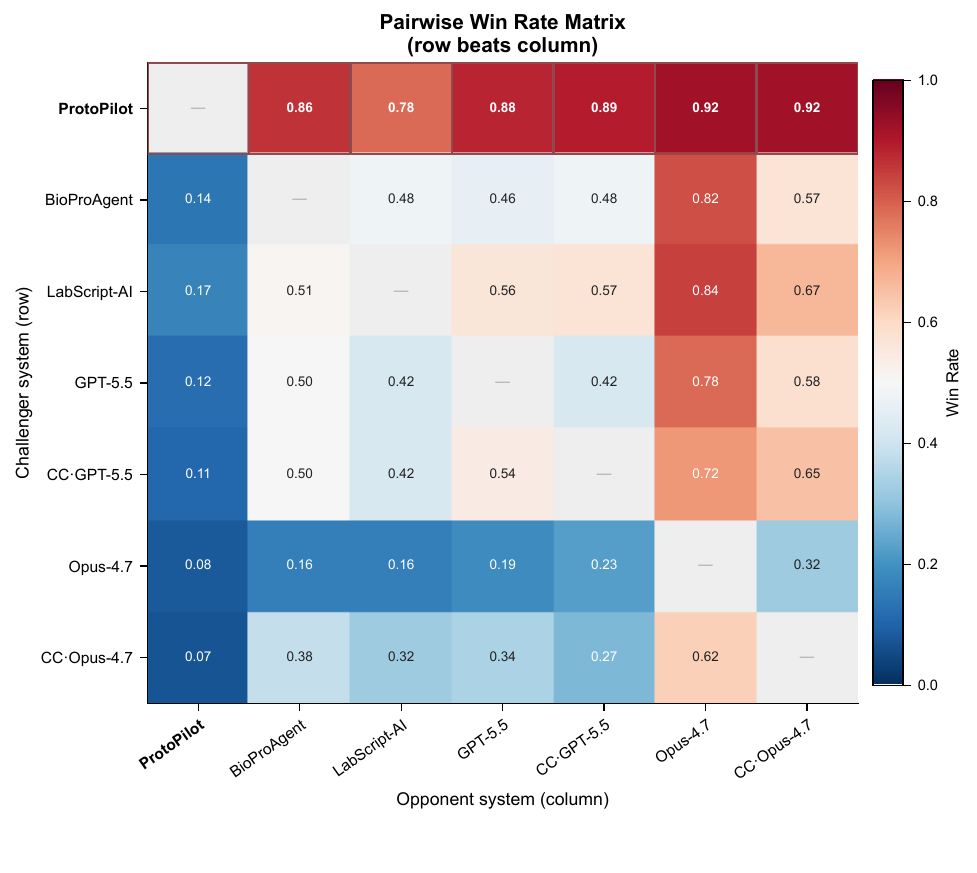}
  \caption{%
    \textbf{Pairwise expert win-rate matrix
    (Supplementary to Fig.~\ref{fig:fig2}d).}
    Each cell $(i, j)$ shows the fraction of (query, expert) pairs for which
    system $i$ was ranked above system $j$.
    Values above 0.50 indicate system $i$ is preferred over system $j$.
    ProtoPilot wins against every other system in the majority of pairwise
    comparisons, with win rates ranging from 0.78 to 0.92.
    $n = 34$ queries $\times$ 3 experts $= 102$ pairwise judgements per cell.
  }
  \label{fig:app_d4}
\end{figure}

\begin{figure}[htbp]
  \centering
  \includegraphics[width=0.95\textwidth]{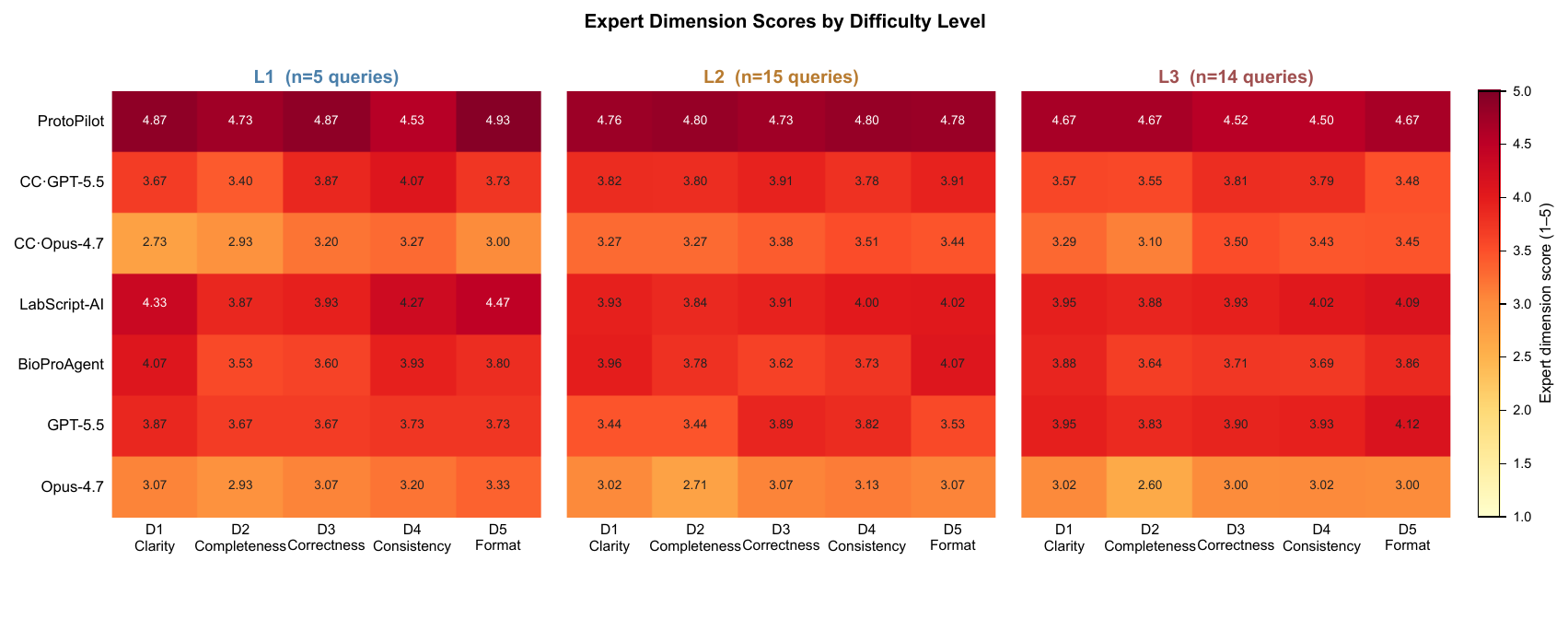}
  \caption{%
    \textbf{Expert dimension scores by difficulty level
    (Supplementary to Fig.~\ref{fig:fig2}d).}
    Heatmap of mean expert scores (1--5 scale) for five evaluation dimensions:
    D1 Clarity, D2 Completeness, D3 Correctness, D4 Consistency, and D5 Format,
    shown separately for L1 ($n = 5$ queries), L2 ($n = 15$), and L3 ($n = 14$).
    ProtoPilot scores consistently near the ceiling of the 5-point scale across
    all dimensions and difficulty levels (range: 4.50--4.93).
  }
  \label{fig:app_d5}
\end{figure}

\begin{figure}[htbp]
  \centering
  \includegraphics[width=0.95\textwidth]{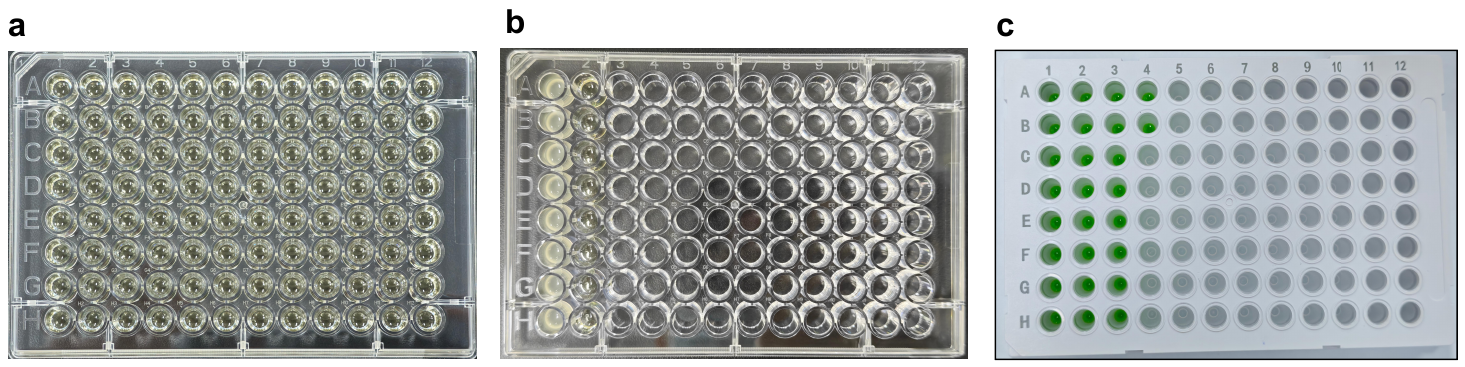}
  \caption{%
    \textbf{Plate results and controls for bacterial inoculation, serial dilution and colony PCR assays (Supplementary to Fig.~\ref{fig:fig3}).}
    \textbf{(a)}~Blank-control plate from the bacterial inoculation assay. Wells remained clear after incubation, indicating no visible contamination in the blank controls.
    \textbf{(b)}~Representative plate image showing the source bacterial culture and diluent blank controls used in the bacterial serial-dilution assay. The source culture wells were visibly turbid, whereas the diluent blank control wells remained clear.
    \textbf{(c)}~Representative dispensing result of the PCR mix in the colony PCR screening assay, showing the reaction setup in the multiwell plate before amplification.
  }
  \label{fig:app_wet3}
\end{figure}

\begin{figure}[htbp]
  \centering
  \includegraphics[width=0.95\textwidth]{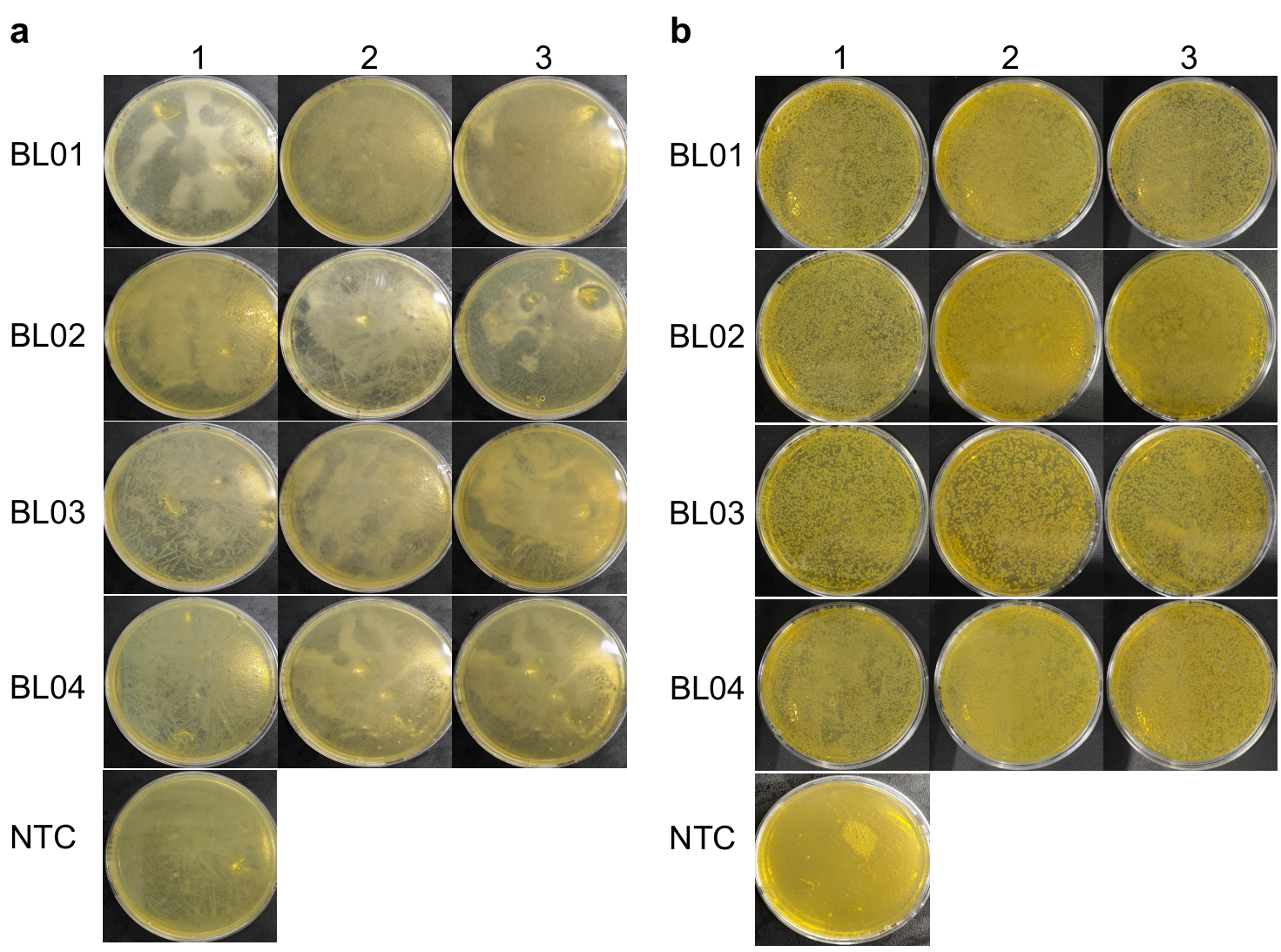}
  \caption{%
    \textbf{All transformation plate results before and after PCA protocol revision (Supplementary to Fig.~\ref{fig:fig6}).}
    All transformation plate images for BL01--BL04 after Gibson assembly.
    \textbf{(a)}~All plates from the first transformation round.
    \textbf{(b)}~All plates from the second transformation round after ProtoPilot-guided failure diagnosis and protocol revision.
    Columns 1--3 denote three transformation replicate plates for each target fragment, and NTC denotes negative control. In the first transformation round, colonies appeared on the NTC plate and the BL01--BL04 sample plates generally showed extensive colony growth or lawn-like growth, making it difficult to pick clear single colonies. After protocol revision, the NTC plate showed no colony growth and the BL01--BL04 sample plates produced single colonies suitable for downstream screening.
  }
  \label{fig:app_wet2}
\end{figure}

\begin{figure}[htbp]
  \centering
  \includegraphics[width=0.95\textwidth]{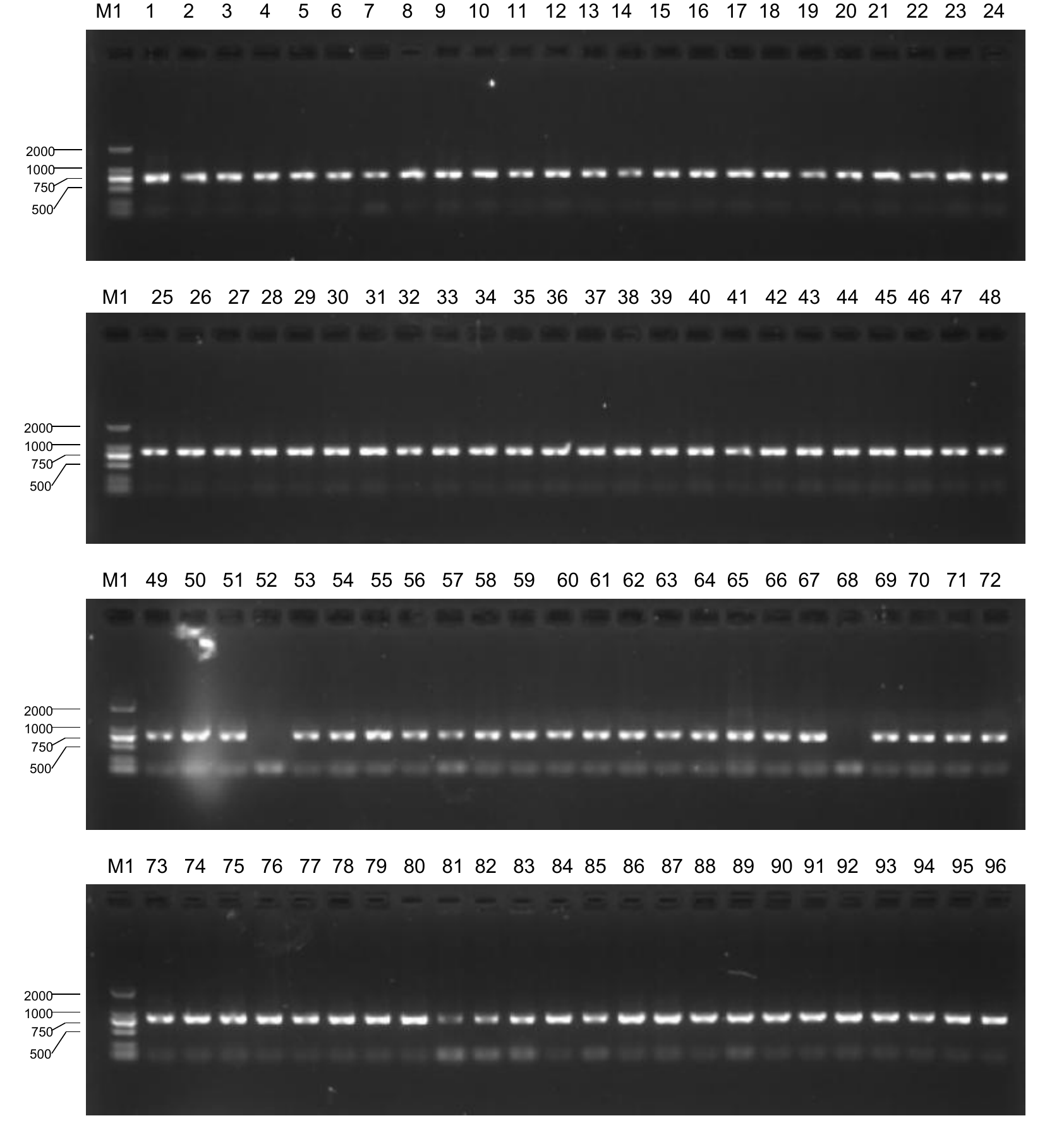}
  \caption{%
    \textbf{Colony PCR screening of PCA-derived clones after transformation (Supplementary to Fig.~\ref{fig:fig6}).}
    Colony PCR screening of BL01--BL04 candidate clones after Gibson assembly and transformation. M1 denotes DL2000 DNA marker, and lanes 1--96 denote 96 candidate clones. Each target fragment was screened using three transformation replicate plates, with eight candidate clones picked from each plate. Lanes 1--24 correspond to BL01 candidates, lanes 25--48 to BL02 candidates, lanes 49--72 to BL03 candidates and lanes 73--96 to BL04 candidates. Clones with clear bands of the expected size were scored as colony-PCR positive; 93 of 96 candidate clones were positive.
  }
  \label{fig:app_wet1}
\end{figure}

\begin{table}[htbp]
\centering
\caption{\textbf{Sanger sequencing confirmation of PCA-derived clones (Supplementary to Fig.~\ref{fig:fig6}).}}
\label{tab:sanger_pca_clones}
\begin{threeparttable}
\footnotesize
\setlength{\tabcolsep}{3.5pt}
\renewcommand{\arraystretch}{1.12}
\begin{tabular}{lccccc}
\toprule
\makecell[l]{Target\\fragment} &
\makecell[c]{Length\\(bp)} &
\makecell[c]{Transformation\\replicate} &
\makecell[c]{Sequenced\\clones} &
\makecell[c]{Sequence-correct\\clones} &
\makecell[c]{Correct clone\\obtained} \\
\midrule
BL01 & 698 & BL01-1 & 4 & 2 & Yes \\
BL01 & 698 & BL01-2 & 4 & 4 & Yes \\
BL01 & 698 & BL01-3 & 4 & 2 & Yes \\
\midrule
BL02 & 802 & BL02-1 & 4 & 2 & Yes \\
BL02 & 802 & BL02-2 & 4 & 2 & Yes \\
BL02 & 802 & BL02-3 & 4 & 2 & Yes \\
\midrule
BL03 & 713 & BL03-1 & 4 & 4 & Yes \\
BL03 & 713 & BL03-2 & 4 & 2 & Yes \\
BL03 & 713 & BL03-3 & 4 & 3 & Yes \\
\midrule
BL04 & 800 & BL04-1 & 4 & 2 & Yes \\
BL04 & 800 & BL04-2 & 4 & 0 & No \\
BL04 & 800 & BL04-3 & 4 & 4 & Yes \\
\bottomrule
\end{tabular}
\begin{tablenotes}
\footnotesize
\item Each target fragment was evaluated across three independent transformation replicates. Four colony-PCR-positive clones were selected for Sanger sequencing from each replicate.
\end{tablenotes}
\end{threeparttable}
\end{table}

\begin{figure}[htbp]
  \centering
  \includegraphics[width=0.95\textwidth]{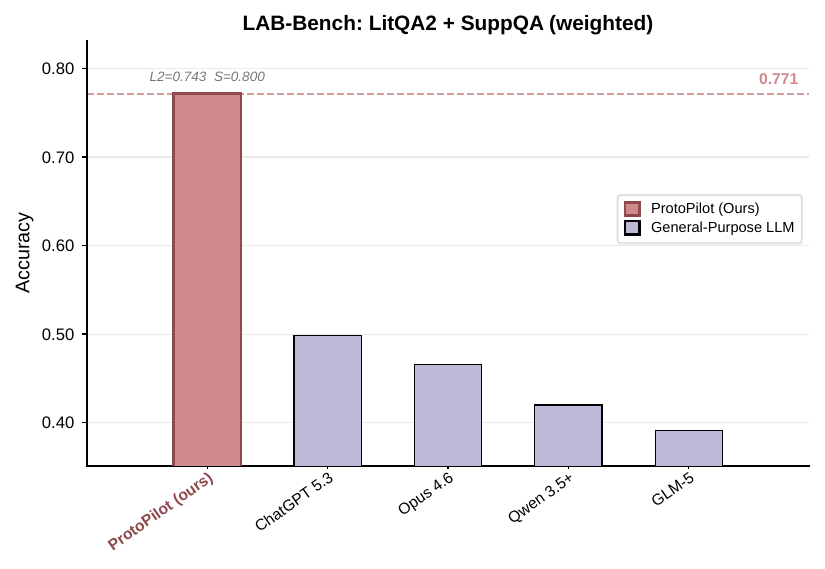}
  \caption{%
     \textbf{LAB-Bench LitQA2 + SuppQA weighted accuracy.}
    Weighted composite score across five systems.
    ProtoPilot achieves 0.772, outperforming ChatGPT~5.3 (0.498), Opus~4.6 (0.466),
    Qwen~3.5+ (0.420), and GLM-5 (0.391).
    Dashed red line indicates ProtoPilot score.
    Individual LitQA2 and SuppQA sub-scores are annotated above each bar where available.
    }
  \label{fig:lab_bench_litqa_suppqa}
\end{figure}

\begin{figure}[htbp]
  \centering
  \includegraphics[width=0.95\textwidth]{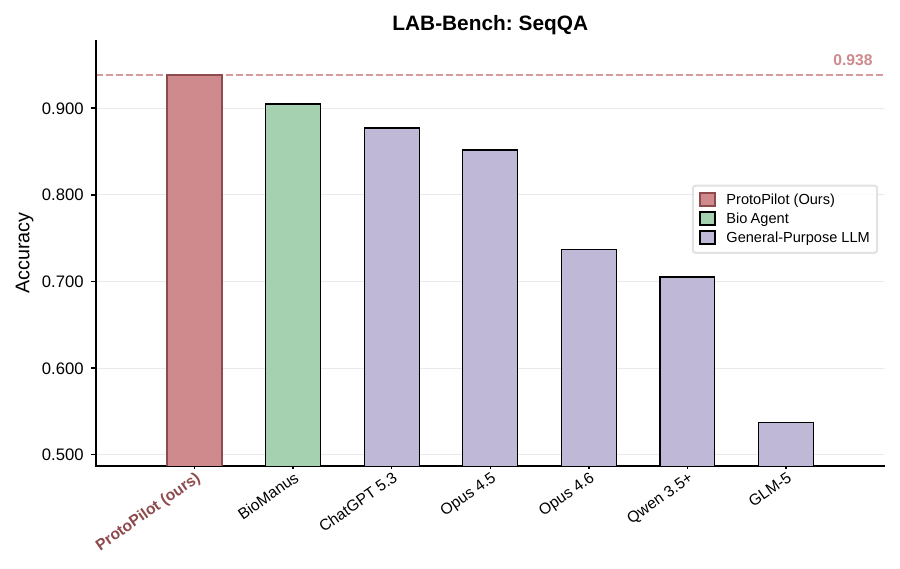}
  \caption{%
     \textbf{LAB-Bench SeqQA accuracy.}
    SeqQA accuracy across seven systems.
    ProtoPilot (0.938) leads BioManus (0.905), ChatGPT~5.3 (0.877), Opus~4.5 (0.852),
    Opus~4.6 (0.737), Qwen~3.5+ (0.705), and GLM-5 (0.537).
    Dashed red line indicates ProtoPilot score.
    }
  \label{fig:lab_bench_seqqa}
\end{figure}

\begin{figure}[htbp]
  \centering
  \includegraphics[width=0.95\textwidth]{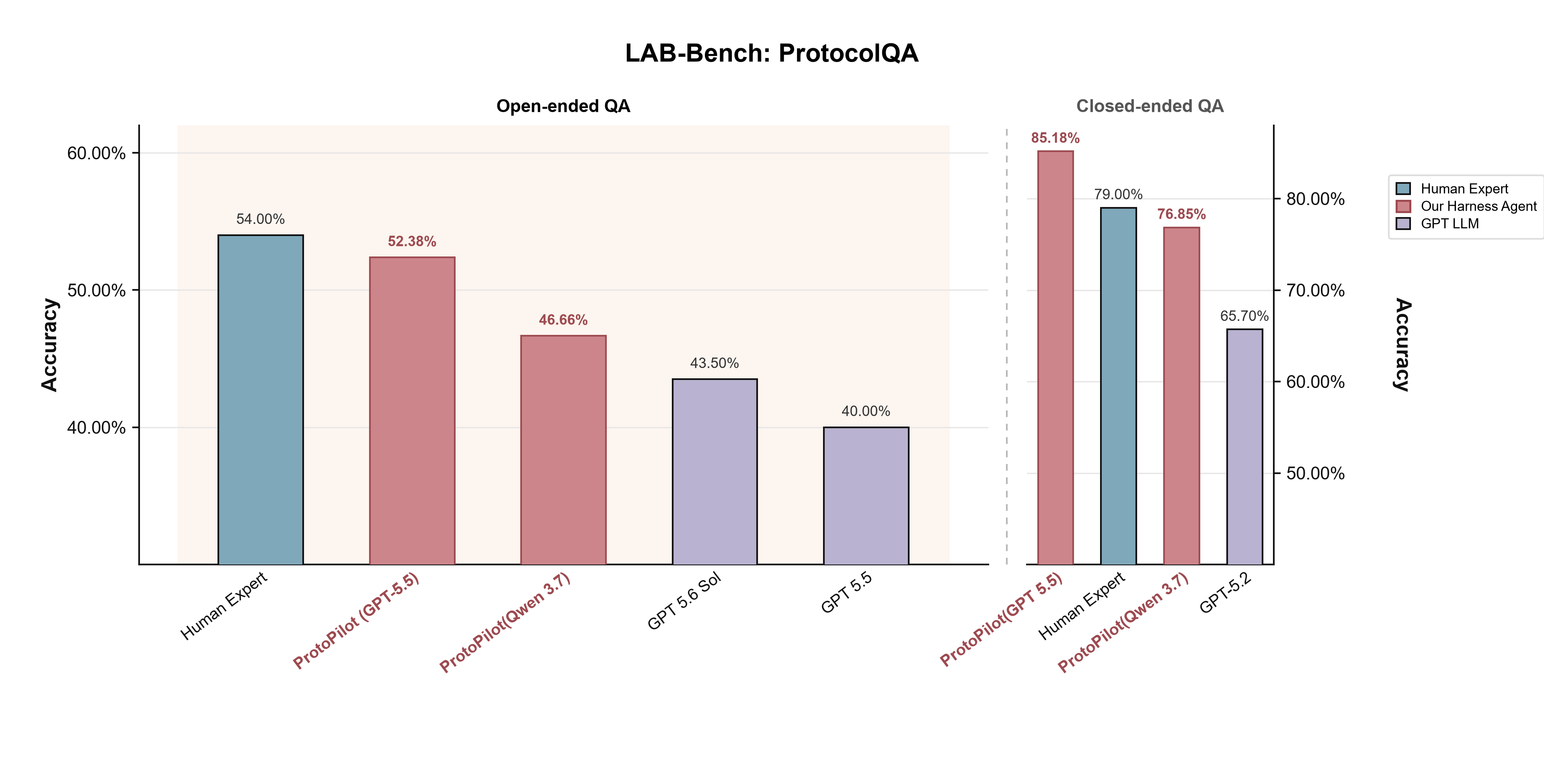}
  \caption{%
     \textbf{LAB-Bench ProtocolQA accuracy.} 
  ProtocolQA accuracy across eight systems.
  ProtoPilot leads all evaluated systems;
    }
  \label{fig:lab_bench_protocolqa}
\end{figure}

\begin{figure}[htbp]
  \centering
  \includegraphics[width=0.95\textwidth]{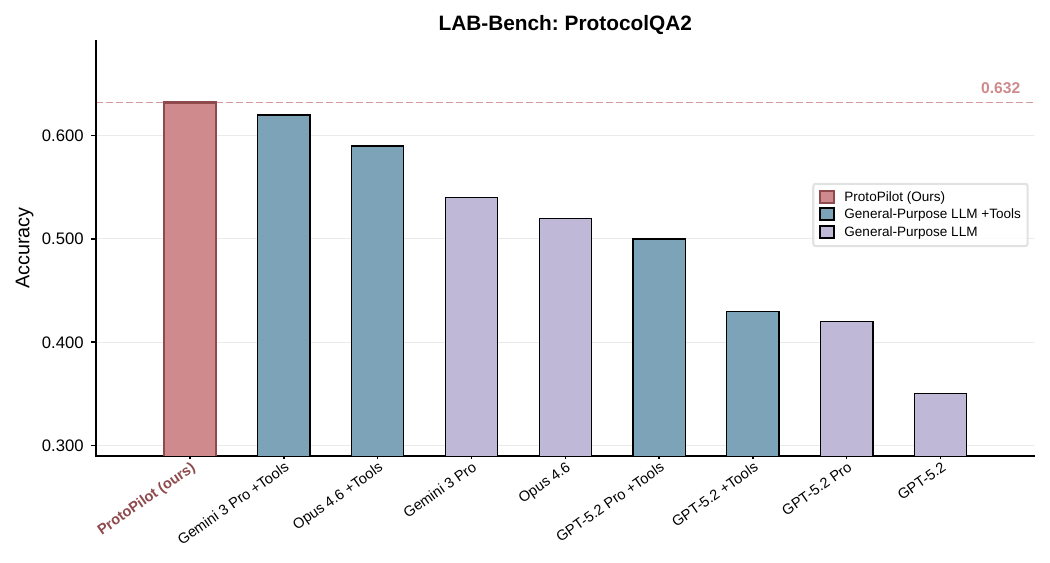}
  \caption{%
     \textbf{LAB-Bench ProtocolQA2 accuracy.}
    ProtocolQA2 accuracy across nine systems, distinguishing pure language models (lavender)
    from tool-use augmented variants (slate blue).
    ProtoPilot (0.632) narrowly leads Gemini~3~Pro~+Tools (0.620) and
    Opus~4.6~+Tools (0.590). Tool-use augmentation systematically improves
    performance over pure-LLM counterparts across all model families.
    Dashed red line indicates ProtoPilot score.
    }
  \label{fig:lab_bench_protocolqa2}
\end{figure}

\end{document}